\documentclass[10pt,twocolumn,letterpaper]{article}

\usepackage{cvpr}
\usepackage{threeparttable}
\usepackage{bbding}
\usepackage[percent]{overpic}
\usepackage[dvipsnames]{xcolor}
\usepackage{times}
\usepackage{epsfig}
\usepackage{graphicx}
\usepackage{amsmath}
\usepackage{amssymb}
\usepackage{mathtools}
\usepackage{multirow}
\usepackage{enumitem}
\usepackage{caption}
\usepackage{subcaption}
\usepackage{array}
\usepackage{colortbl}
\usepackage{booktabs}
\usepackage{bbm}
\usepackage{multicol}
\usepackage{makecell}
\usepackage{xspace}
\usepackage{float}
\usepackage{color}
\usepackage{lipsum}
\usepackage{pifont}
\usepackage{cite}
\usepackage[pagebackref=true,breaklinks=true,letterpaper=true,colorlinks,bookmarks=false]{hyperref}
\usepackage[capitalize]{cleveref}

\usepackage{array}
\newcolumntype{H}{>{\setbox0=\hbox\bgroup}c<{\egroup}@{}}

\cvprfinalcopy % *** Uncomment this line for the final submission

\def\cvprPaperID{****} % *** Enter the CVPR Paper ID here

\newcommand\blfootnote[1]{%
  \begingroup
  \renewcommand\thefootnote{}\footnote{#1}%
  \addtocounter{footnote}{-1}%
  \endgroup
}
\begin{document}

%%%%%%%%% TITLE
\title{YOLOv6: A Single-Stage Object Detection Framework for Industrial Applications}

\author{Chuyi Li$^*$~~~ Lulu Li$^*$~~~ Hongliang Jiang$^*$~~~ Kaiheng Weng$^*$~~~ Yifei Geng$^*$~~~ Liang Li$^*$\\~~~ Zaidan Ke$^*$ ~~~ 
Qingyuan Li$^*$~~~ Meng Cheng$^*$~~~ Weiqiang Nie$^*$~~~ Yiduo Li$^*$~~~ Bo Zhang$^*$~~~ \\ Yufei Liang~~~ Linyuan Zhou~~~  Xiaoming Xu$^\dag$~~~ Xiangxiang Chu~~~ Xiaoming Wei~~~  Xiaolin Wei\\
	Meituan Inc.\\
	\tt\small \{lichuyi, lilulu05, jianghongliang02, wengkaiheng, gengyifei,liliang58, kezaidan,\\
  \tt\small  liqingyuan02, chengmeng05, nieweiqiang, liyiduo, zhangbo97, liangyufei, zhoulinyuan, \\
  \tt\small  xuxiaoming04, chuxiangxiang, weixiaoming, weixiaolin02\}@meituan.com\\ 
    }

\twocolumn[
{\renewcommand\twocolumn[1][]{#1}
\maketitle
\vspace{-11mm}
\begin{figure}[H]
\hsize=\textwidth
\centering
\begin{subfigure}{0.48\textwidth}
\centering
\includegraphics[width=1\textwidth]{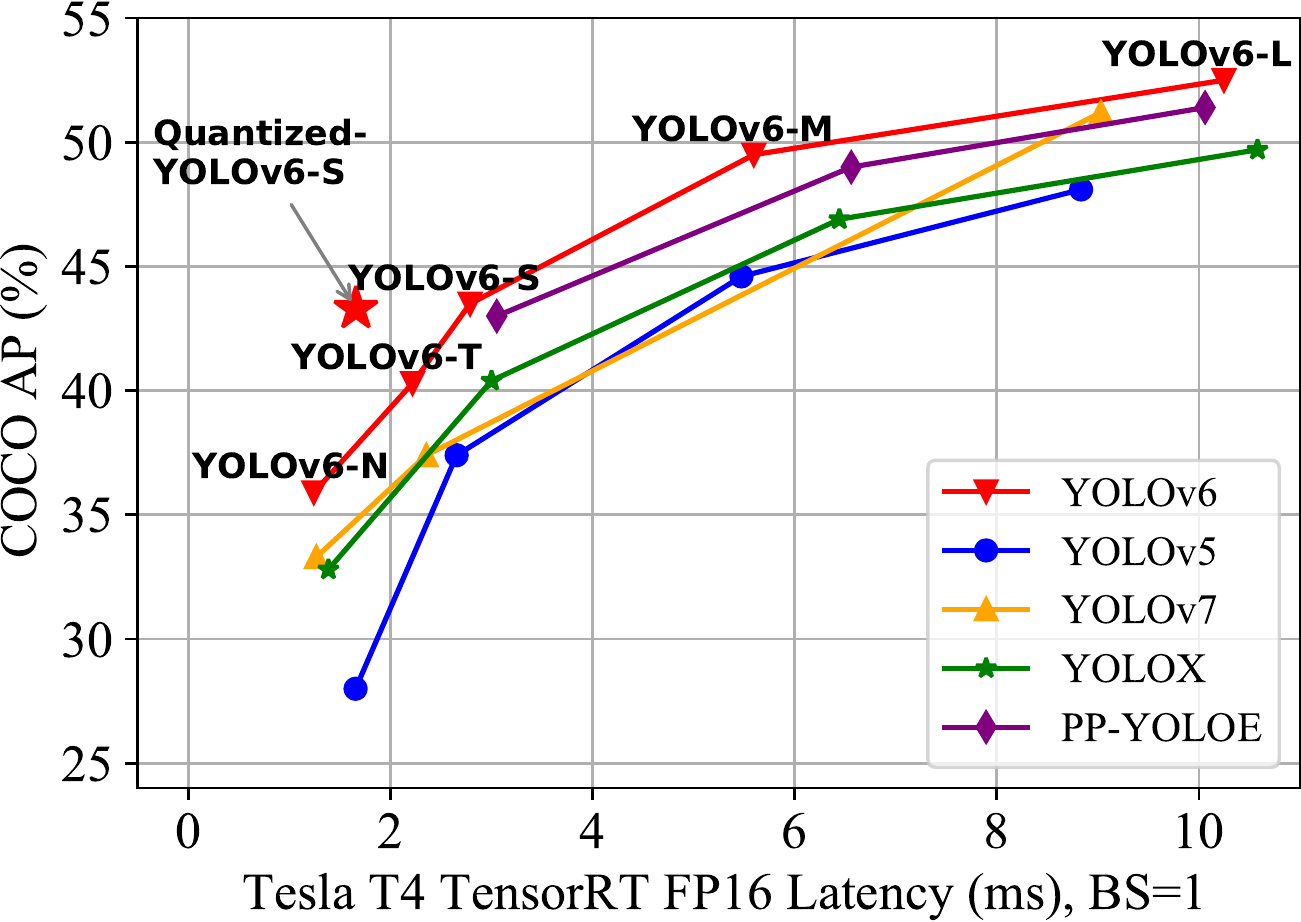}
%\caption{aaaaa}
\label{fig:1a}	
\end{subfigure}    
\hspace{0.2in}
\begin{subfigure}{0.48\textwidth}
\centering
\includegraphics[width=1\textwidth]{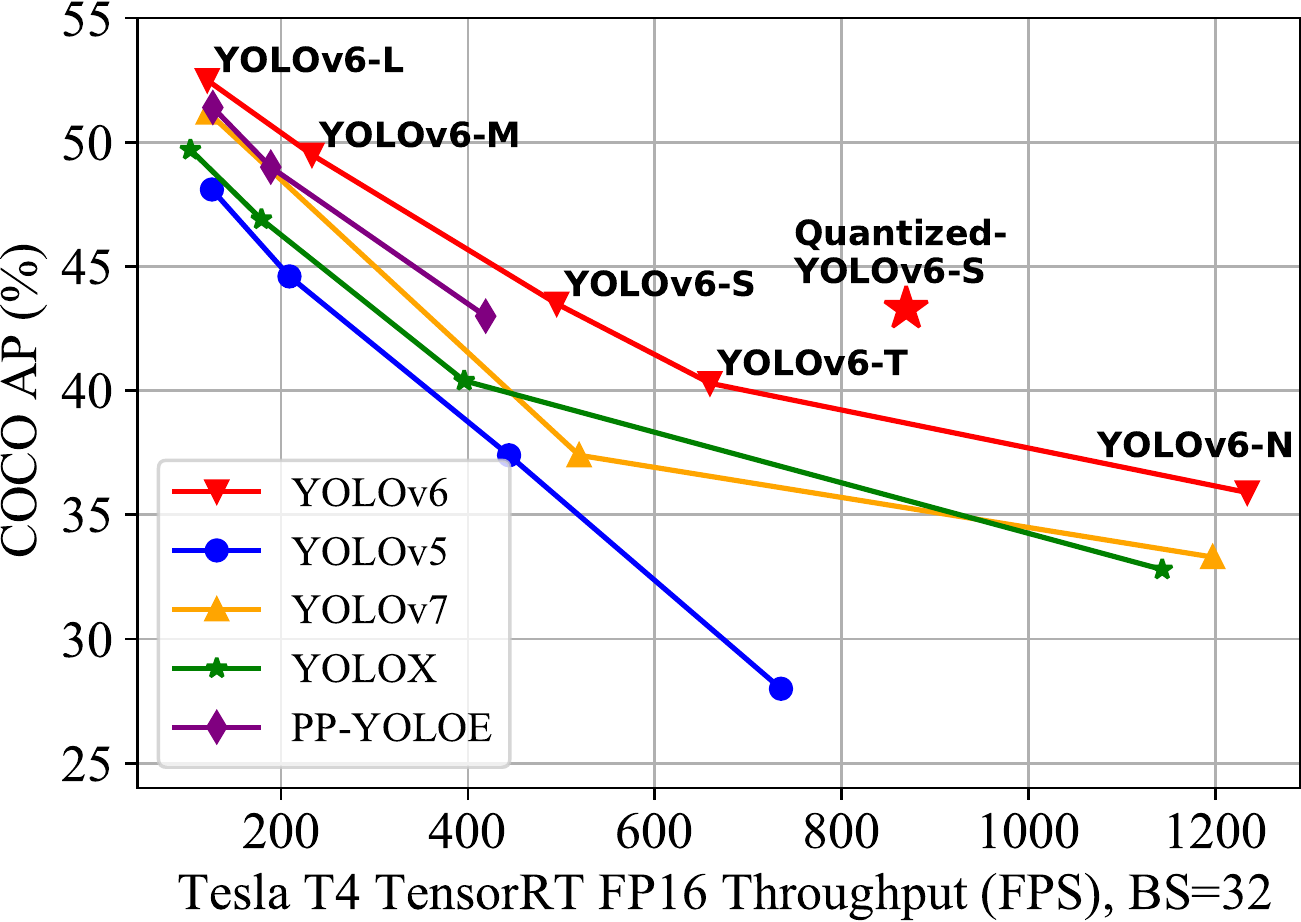}
%\caption{bbbbb}
\label{fig:1b}
\end{subfigure}
\hspace{0.in}
\vspace{-6mm}
\caption{Comparison of state-of-the-art efficient object detectors. Both latency and throughput (at a batch size of 32) are given for a handy reference. All models are test with TensorRT 7 except that the quantized model is with TensorRT 8.}
\label{fig:sota-comp}
\end{figure}
% \vspace{2mm}
}
]

\blfootnote{* Equal contributions.}
\blfootnote{\dag~Corresponding author.}
%%%%%%%%% ABSTRACT

\begin{abstract}
  For years, YOLO series have been de facto industry-level standard for efficient object detection. The YOLO community has prospered overwhelmingly to enrich its use in a multitude of hardware platforms and abundant scenarios. In this technical report, we strive to push its limits to the next level, stepping forward with an unwavering mindset for industry application.
  Considering the diverse requirements for speed and accuracy in the real environment, we extensively examine the up-to-date object detection advancements either from industry or academy. Specifically, we heavily assimilate ideas from recent network design, training strategies, testing techniques, quantization and optimization methods. On top of this, we integrate our thoughts and practice to build a suite of deployment-ready networks at various scales to accommodate diversified use cases. With the generous permission of YOLO authors, we name it YOLOv6. We also express our warm welcome to users and contributors for further enhancement. For a glimpse of performance, our YOLOv6-N hits 35.9\% AP on COCO dataset at a throughput of 1234 FPS on an NVIDIA Tesla T4 GPU. YOLOv6-S strikes 43.5\% AP at 495 FPS, outperforming other mainstream detectors at the same scale~(YOLOv5-S, YOLOX-S and PPYOLOE-S). Our quantized version of YOLOv6-S even brings a new state-of-the-art 43.3\% AP at 869 FPS. Furthermore, YOLOv6-M/L also achieves better accuracy performance (i.e., 49.5\%/52.3\%) than other detectors with the similar inference speed. We carefully conducted experiments to validate the effectiveness of each component. Our code is made available at~\url{https://github.com/meituan/YOLOv6}.
\end{abstract}

%%%%%%%%% BODY TEXT

\section{Introduction}
\label{sec:intro}
YOLO series have been the most popular detection frameworks in industrial applications, for its excellent balance between speed and accuracy. Pioneering works of YOLO series are YOLOv1-3~\cite{redmon2016you,redmon2017yolo9000,redmon2018yolov3}, which blaze a new trail of one-stage detectors along with the later substantial improvements. YOLOv4~\cite{bochkovskiy2020yolov4} reorganized the detection framework into several separate parts (backbone, neck and head), and verified bag-of-freebies and bag-of-specials at the time to design a framework suitable for training on  a single GPU. At present, YOLOv5~\cite{yolov5}, YOLOX~\cite{ge2021yolox}, PPYOLOE~\cite{xu2022pp} and YOLOv7~\cite{wang2022yolov7} are all the competing candidates for efficient detectors to deploy. Models at different sizes are commonly obtained  through scaling techniques.

% \begin{figure}[t]
% 	\centering
% 	\includegraphics[width=8cm]{fig/yolo-comparison-latency.pdf}
% 	\caption{Comparison of state-of-the-art efficient object detectors.}
% 	\label{fig:sota-comp}
% \end{figure}

In this report, we empirically observed several important factors that motivate us to refurnish the YOLO framework: \textbf{(1)} Reparameterization from RepVGG~\cite{ding2021repvgg} is a superior technique that is not yet well exploited in detection. We also notice that simple model scaling for RepVGG blocks becomes impractical, for which we consider that the elegant consistency of the network design between small and large networks is unnecessary. The plain single-path architecture is a better choice for small networks, but for larger models, the exponential growth of the parameters and the computation cost of the single-path architecture makes it infeasible; \textbf{(2)} Quantization of reparameterization-based detectors also requires meticulous treatment, otherwise it would be intractable to deal with performance degradation due to its heterogeneous configuration during training and inference.  \textbf{(3)} Previous works \cite{yolov5,xu2022pp,ge2021yolox,wang2022yolov7} tend to pay less attention to deployment, whose latencies are commonly compared on high-cost machines like V100. There is a hardware  gap when it comes to real serving environment. Typically, low-power GPUs like Tesla T4 are less costly and provide rather good inference performance.  \textbf{(4)} Advanced domain-specific strategies like label assignment and loss function design need further verifications considering the architectural variance;
\textbf{(5)} For deployment, we can tolerate the adjustments of the training strategy that improve the accuracy performance but not increase inference costs, such as \emph{knowledge distillation}.

\begin{figure*}[t]
  \begin{center}
    \includegraphics[width=0.99\linewidth]{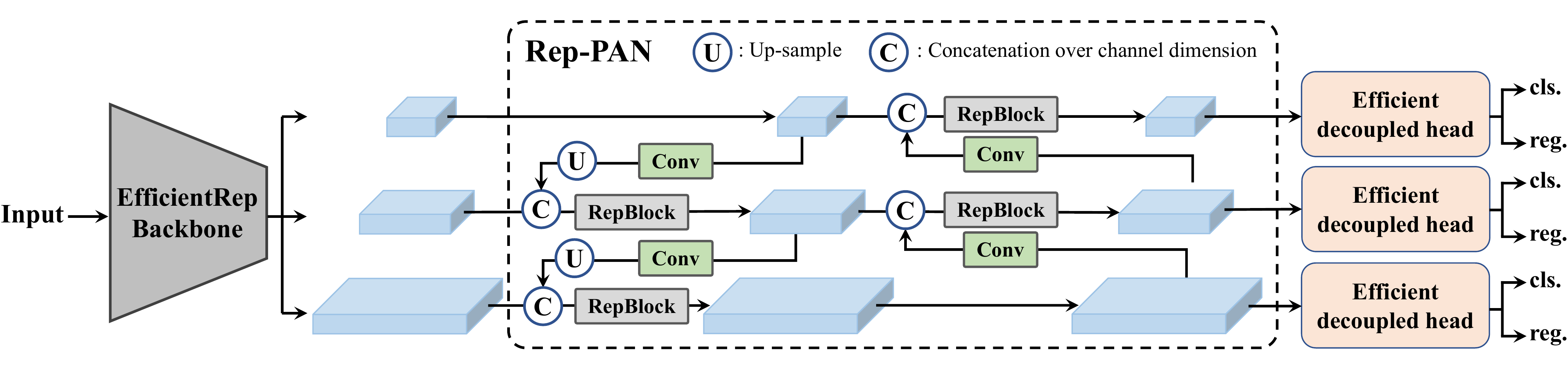}
  \end{center}
    \caption{The YOLOv6 framework (N and S are shown). Note for M/L, RepBlocks is replaced with CSPStackRep.}
    \label{fig:framework}
\end{figure*}

%CSP connection~\cite{wang2020cspnet} and the residual connection~\cite{he2016deep} could balance the speed and the accuracy better compared to the single-path architecture. Likewise, the selection of the activation functions and the convolution layer could vary considering the discrepancy of the networks' representation ability. Another example is the introduction of probability loss~\cite{li2020generalized,li2021generalized} in the box regression. It brings extra parameters, which have marginal effects on the speed performance of larger models but remarkably influence small models; 

With the aforementioned observations in mind, we bring the birth of YOLOv6, which accomplishes so far the best trade-off in terms of accuracy and speed. We show the comparison of YOLOv6 with other peers at a similar scale in~\cref{fig:sota-comp}. To boost inference speed without much performance degradation, we examined the cutting-edge quantization methods, including post-training quantization (PTQ) and quantization-aware training (QAT), and accommodate them in YOLOv6 to achieve the goal of deployment-ready networks.

We summarize the main aspects of YOLOv6 as follows:
\begin{itemize}
\item We refashion a line of networks of different sizes tailored for industrial applications in diverse scenarios. The architectures at different scales vary to achieve the best speed and accuracy trade-off, where small models feature a plain single-path backbone and large models are built on efficient multi-branch blocks.
\item We imbue YOLOv6 with a self-distillation strategy, performed both on the classification task and the regression task. Meanwhile, we dynamically adjust the knowledge from the teacher and labels to help the student model learn knowledge more efficiently during all training phases.
\item We broadly verify the advanced detection techniques for label assignment, loss function and data augmentation techniques and adopt them selectively to further boost the performance.
\item We reform the quantization scheme for detection with the help of RepOptimizer~\cite{ding2022re} and channel-wise distillation~\cite{shu2021channel}, which leads to an ever-fast and accurate detector with 43.3\% COCO AP and a throughput of 869 FPS at a batch size of 32. 
\end{itemize}

\section{Method}
\label{sec:method}
The renovated design of YOLOv6 consists of the following components, \emph{network design}, \emph{label assignment}, \emph{loss function}, \emph{data augmentation}, \emph{industry-handy improvements}, and \emph{quantization and deployment}:
\begin{itemize}
  \item \textbf{Network Design:}
  \emph{Backbone}: Compared with other mainstream architectures, we find that RepVGG~\cite{ding2021repvgg} backbones are equipped with more feature representation power in small networks at a similar inference speed, whereas it can hardly be scaled to obtain larger models due to the explosive growth of the parameters and computational costs. In this regard, we take RepBlock~\cite{ding2021repvgg} as the building block of our small networks. For large models, we revise a more efficient CSP\cite{wang2020cspnet} block, named \emph{CSPStackRep Block}.
  \emph{Neck}: The neck of YOLOv6 adopts PAN topology~\cite{liu2018path} following YOLOv4 and YOLOv5. We enhance the neck with RepBlocks or CSPStackRep Blocks to have Rep-PAN.
  \emph{Head}: We simplify the decoupled head to make it more efficient, called \emph{Efficient Decoupled Head}.
 
  \item \textbf{Label Assignment:} We evaluate the recent progress of label assignment strategies ~\cite{zhang2020atss,ge2021yolox,feng2021tood,dw_CVPR,Zand2022ObjectBoxFC} on YOLOv6 through numerous experiments, and the results indicate that TAL~\cite{feng2021tood} is more effective and training-friendly.
  \item \textbf{Loss Function:} The loss functions of the mainstream anchor-free object detectors contain \emph{classification loss}, \emph{box regression loss} and \emph{object loss}. For each loss, we systematically experiment it with all available techniques and finally select VariFocal Loss~\cite{zhang2021varifocalnet} as our classification loss and SIoU~\cite{gevorgyan2022siou}/GIoU~\cite{rezatofighi2019generalized} Loss as our regression loss. 
  % Note that we deprecate object loss for its negative effect.
  % \item \textbf{Data Augmentation:} Mosaic augmentation~\cite{bochkovskiy2020yolov4} and Mixup augmentation~\cite{zhang2017mixup} are two prevailing strong augmentations in object detection. In YOLOv6, these two augmentations are also adopted. However, there are some newly proposed tricks to modify these two strategies in YOLOX~\cite{ge2021yolox}, which are deserved systematic verification. As a result, we searched for the best augmentation combinations for all YOLOv6 networks. 
  \item \textbf{Industry-handy improvements:} We introduce additional common practice and tricks to improve the performance including \emph{self-distillation} and \emph{more training epochs}. For self-distillation, both classification and box regression are respectively supervised by the teacher model. The distillation of box regression is made possible thanks to DFL~\cite{li2020generalized}. In addition, the proportion of information from the soft and hard labels is dynamically declined via cosine decay, which helps the student selectively acquire knowledge at different phases during the training process. In addition, we encounter the problem of the impaired performance without adding extra gray borders at evaluation, for which we provide some remedies.
  \item \textbf{Quantization and deployment:} To cure the performance degradation in quantizing reparameterization-based models, we train YOLOv6 with RepOptimizer~\cite{ding2022re} to obtain PTQ-friendly weights. We further adopt QAT with channel-wise distillation~\cite{shu2021channel} and graph optimization to pursue extreme performance. Our quantized YOLOv6-S hits a new state of the art with 42.3\% AP and a throughput of 869 FPS (batch size=32).
  \end{itemize}

\subsection{Network Design}
  \label{sec:method:network}
  A one-stage object detector is generally composed of the following parts: a backbone, a neck and a head. The backbone mainly determines the feature representation ability, meanwhile, its design has a critical influence on the inference efficiency since it carries a large portion of computation cost. The neck is used to aggregate the low-level physical features with high-level semantic features, and then build up pyramid feature maps at all levels. The head consists of several convolutional layers, and it predicts final detection results according to multi-level features assembled by the neck. It can be categorized as \emph{anchor-based} and \emph{anchor-free}, or rather \emph{parameter-coupled head} and \emph{parameter-decoupled head} from the structure's perspective.

  In YOLOv6, based on the principle of hardware-friendly network design~\cite{ding2021repvgg}, we propose two scaled re-parameterizable backbones and necks to accommodate models at different sizes, as well as an efficient decoupled head with the hybrid-channel strategy. The overall architecture of YOLOv6 is shown in~\cref{fig:framework}. 

\subsubsection{Backbone}
\label{sec:method:backbone}
As mentioned above, the design of the backbone network has a great impact on the effectiveness and efficiency of the detection model. Previously, it has been shown that multi-branch networks~\cite{szegedy2015going,szegedy2016rethinking,he2016identity,huang2017densely} can often achieve better classification performance than single-path ones~\cite{krizhevsky2012imagenet, simonyan2014very}, but often it comes with the reduction of the parallelism and results in an increase of inference latency. On the contrary, plain single-path networks like VGG~\cite{simonyan2014very} take the advantages of high parallelism and less memory footprint, leading to higher inference efficiency. Lately in RepVGG~\cite{ding2021repvgg}, a structural re-parameterization method is proposed to decouple the training-time multi-branch topology with an inference-time plain architecture to achieve a better speed-accuracy trade-off. 

\begin{figure}[htp]
  \centering
  \includegraphics[width=\columnwidth]{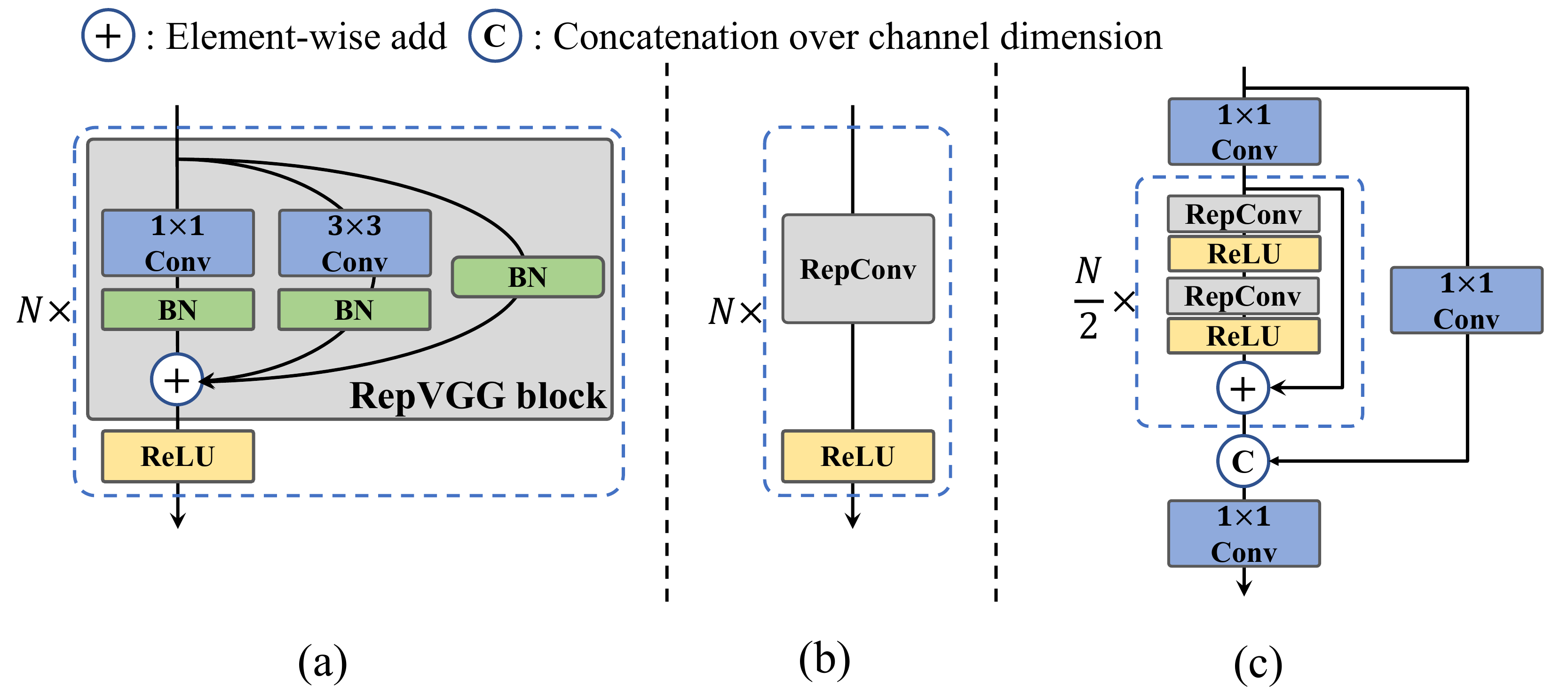}
  \caption{(a) RepBlock is composed of a stack of RepVGG blocks with ReLU activations at training. (b) During inference time, RepVGG block is converted to RepConv. (c) CSPStackRep Block comprises three 1$\times$1 convolutional layers and a stack of sub-blocks of double RepConvs following the ReLU activations with a residual connection. }
  \label{fig:repblock}
\end{figure}

Inspired by the above works, we design an efficient re-parameterizable backbone denoted as \emph{EfficientRep}. For small models, the main component of the backbone is RepBlock during the training phase, as shown in~\cref{fig:repblock} (a). And each RepBlock is converted to stacks of 3 $\times$ 3 convolutional layers (denoted as RepConv) with ReLU activation functions during the inference phase, as shown in~\cref{fig:repblock} (b). Typically a 3$\times$3 convolution is highly optimized on mainstream GPUs and CPUs and it enjoys higher computational density. Consequently, EfficientRep Backbone sufficiently utilizes the computing power of the hardware, resulting in a significant decrease in inference latency while enhancing the representation ability in the meantime.

However, we notice that with the model capacity further expanded, the computation cost and the number of parameters in the single-path plain network grow exponentially. To achieve a better trade-off between the computation burden and accuracy, we revise a CSPStackRep Block to build the backbone of medium and large networks. As shown in~\cref{fig:repblock} (c), CSPStackRep Block is composed of three 1$\times$1 convolution layers and a stack of sub-blocks consisting of two RepVGG blocks~\cite{ding2021repvgg} or RepConv (at training or inference respectively) with a residual connection. Besides, a \emph{cross stage partial} (CSP) connection is adopted to boost performance without excessive computation cost. Compared with CSPRepResStage \cite{xu2022ppyoloe}, it comes with a more succinct outlook and considers the balance between accuracy and speed.

\subsubsection{Neck}
In practice, the feature integration at multiple scales has been proved to be a critical and effective part of object detection~\cite{lin2017feature, liu2018path, tan2020efficientdet, ghiasi2019fpn}. We adopt the modified PAN topology~\cite{liu2018path} from YOLOv4~\cite{bochkovskiy2020yolov4} and YOLOv5~\cite{yolov5} as the base of our detection neck. In addition, we replace the CSP-Block used in YOLOv5 with RepBlock (for small models) or CSPStackRep Block (for large models) and adjust the width and depth accordingly. The neck of YOLOv6 is denoted as Rep-PAN.

\subsubsection{Head}
  \paragraph{Efficient decoupled head} 
  The detection head of YOLOv5 is a coupled head with parameters shared between the classification and localization branches, while its counterparts in FCOS~\cite{tian2019fcos} and YOLOX~\cite{ge2021yolox} decouple the two branches, and additional two 3$\times$3 convolutional layers are introduced in each branch to boost the performance. 

  In YOLOv6, we adopt a \emph{hybrid-channel} strategy to build a more efficient decoupled head. Specifically, we reduce the number of the middle 3$\times$3 convolutional layers to only one. The width of the head is jointly scaled by the width multiplier for the backbone and the neck. These modifications further reduce computation costs to achieve a lower inference latency.

  \paragraph{Anchor-free}
  % Since some anchor-based detectors~\cite{redmon2018yolov3,bochkovskiy2020yolov4} need cluster analysis before training to discover the best anchor settings, it will increase the complexity of the detector and the difficulties of detecting objects of extreme scales and aspect ratios to a certain extent. Besides, the additional time usage caused by data transferring of prediction results between different hardware in some industrial applications on edge could not be ignored.

  Anchor-free detectors stand out because of their better generalization ability and simplicity in decoding prediction results. The time cost of its post-processing is substantially reduced. There are two types of anchor-free detectors: anchor point-based~\cite{tian2019fcos,ge2021yolox} and keypoint-based~\cite{zhou2019objects,law2018cornernet,yang2019reppoints}. In YOLOv6, we adopt the anchor point-based paradigm, whose box regression branch actually predicts the distance from the anchor point to the four sides of the bounding boxes.
  
  % In recent years, the anchor-free paradigms of anchor-point based~\cite{tian2019fcos,ge2021yolox} and keypoint-based~\cite{zhou2019objects,law2018cornernet,yang2019reppoints} have been widely used because of better generalization abilities and simpler logic for decoding prediction results. In YOLOv6, we adopt the anchor-point based paradigm, which actually predicts the distance from the anchor-point to the four sides of the bounding boxes. 
  % We further experimentally verify that the anchor-free paradigm we use could benefit from the 3 times less number of dimensionality in predict results when compared with anchor-based paradigm in YOLOv5, and result in obtaining a 51\% improvement in speed as shown in Table 4.

  % \begin{figure}[htp]
  %   \centering
  %   \includegraphics[width=\columnwidth]{fig/yolov6-head.pdf}
  %   \caption{YOLOv6's Efficient Decoupled Head.}
  %   \label{fig:head}
  % \end{figure}

  \subsection{Label Assignment}
  \label{sec:labelassign}
  Label assignment is responsible for assigning labels to predefined anchors during the training stage. Previous work has proposed various label assignment strategies ranging from simple IoU-based strategy and inside ground-truth method~\cite{tian2019fcos} to other more complex schemes~\cite{zhang2020atss,ge2021yolox,feng2021tood,dw_CVPR,Zand2022ObjectBoxFC}.
  
  \paragraph{SimOTA}
  OTA~\cite{Ge2021OTAOT} considers the label assignment in object detection as an optimal transmission problem. It defines positive/negative training samples for each ground-truth object from a global perspective. SimOTA~\cite{ge2021yolox} is a simplified version of OTA~\cite{Ge2021OTAOT}, which reduces additional hyperparameters and maintains the performance. SimOTA was utilized as the label assignment method in the early version of YOLOv6. However, in practice, we find that introducing SimOTA will slow down the training process. And it is not rare to fall into unstable training. Therefore, we desire a replacement for SimOTA.

  \paragraph{Task alignment learning}
  Task Alignment Learning (TAL) was first proposed in TOOD~\cite{feng2021tood}, in which a unified metric of classification score and predicted box quality is designed. The IoU is replaced by this metric to assign object labels. To a certain extent, the problem of the misalignment of tasks (classification and box regression) is alleviated.
 
 The other main contribution of TOOD is about the task-aligned head (T-head). T-head stacks convolutional layers to build interactive features, on top of which the Task-Aligned Predictor (TAP) is used. PP-YOLOE~\cite{xu2022ppyoloe} improved T-head by replacing the layer attention in T-head with the lightweight ESE attention, forming ET-head. However, we find that the ET-head will deteriorate the inference speed in our models and it comes with no accuracy gain. Therefore, we retain the design of our Efficient decoupled head.
 
 Furthermore, we observed that TAL could bring more performance improvement than SimOTA and stabilize the training. Therefore, we adopt TAL as our default label assignment strategy in YOLOv6.
  % TAL does two jobs, label assign and re-weighting loss. The authors specifically design a unified measure of classification score and predicted box quality. Now, the basis of label assignment is no longer the IOU, but the representation of joint classification and localization. To a certain extent, the problem of misalignment of decoupling head tasks is alleviated. In addition, it also makes NMS retain low-scoring high-quality boxes when sorting prediction boxes, and discard high-scoring low-quality boxes. Due to these characteristics, TAL benefit the detection algorithm and also proves very effective in our YOLOv6.
  
  % \paragraph{DW \& Objectbox} DW~\cite{dw_CVPR} and ObjectBox~\cite{Zand2022ObjectBoxFC} are additional two recent label assignment methods.
  % \paragraph{Others}
  % We spend lots of effort on experimenting with different label assignment methods on YOLOv6 since most of them add no extra computation cost during inference. As shown in~\cref{sec:exp:ablate:labelassign}, there are several additional recent label assignment methods~\cite{zhang2020atss,dw_CVPR,Zand2022ObjectBoxFC} have been experimented. As a result, TAL is adopted in YOLOv6 for its highest accuracy gain.
  
  \subsection{Loss Functions}
  Object detection contains two sub-tasks: \emph{classification} and \emph{localization}, corresponding to two loss functions: \emph{classification loss} and \emph{box regression loss}. For each sub-task, there are various loss functions presented in recent years. In this section, we will introduce these loss functions and describe how we select the best ones for YOLOv6.
  % Loss is calculated on the prediction head, and the structure of the prediction head determines the choice of loss. The head of a detection algorithm usually consists of several branches, a classification branch, a regression branch and an optional object branch, as shown in~\cref{fig:loss}. In this section we will introduce these losses and describe how to choose the best ones for YOLOv6.

  % \begin{figure}[htp]
  %     \centering
  %     \includegraphics[width=8cm]{fig/loss.png}
  %     \caption{Loss}
  %     \label{fig:loss}
  % \end{figure}
  
  \subsubsection{Classification Loss}
  Improving the performance of the classifier is a crucial part of optimizing detectors. Focal Loss~\cite{lin2017focal} modified the traditional cross-entropy loss to solve the problems of class imbalance either between positive and negative examples, or hard and easy samples. To tackle the inconsistent usage of the quality estimation and classification between training and inference, Quality Focal Loss (QFL)~\cite{li2020generalized} further extended Focal Loss with a joint representation of the classification score and the localization quality for the supervision in classification. Whereas VariFocal Loss (VFL)~\cite{zhang2021varifocalnet} is rooted from Focal Loss~\cite{lin2017focal}, but it treats the positive and negative samples asymmetrically. By considering positive and negative samples at different degrees of importance, it balances learning signals from both samples.
  Poly Loss~\cite{leng2022polyloss} decomposes the commonly used classification loss into a series of weighted polynomial bases. It tunes polynomial coefficients on different tasks and datasets, which is proved better than Cross-entropy Loss and Focal Loss through experiments.
  
  We assess all these advanced classification losses on YOLOv6 to finally adopt VFL~\cite{zhang2021varifocalnet}.
  
  \subsubsection{Box Regression Loss}
  Box regression loss provides significant learning signals localizing bounding boxes precisely. L1 loss is the original box regression loss in early works. Progressively, a variety of well-designed box regression losses have sprung up, such as IoU-series loss~\cite{yu2016unitbox, zheng2020distance,rezatofighi2019generalized, zheng2020distance,he2021alpha,gevorgyan2022siou} and probability loss~\cite{li2020generalized}.

  \paragraph{IoU-series Loss}
  IoU loss~\cite{yu2016unitbox} regresses the four bounds of a predicted box as a whole unit. It has been proved to be effective because of its consistency with the evaluation metric. There are many variants of IoU, such as GIoU~\cite{rezatofighi2019generalized}, DIoU~\cite{zheng2020distance}, CIoU~\cite{zheng2020distance}, $\alpha$-IoU~\cite{he2021alpha} and SIoU~\cite{gevorgyan2022siou}, etc, forming relevant loss functions. We experiment with GIoU, CIoU and SIoU in this work. And SIoU is applied to YOLOv6-N and YOLOv6-T, while others use GIoU.
  
  \paragraph{Probability Loss}
  Distribution Focal Loss (DFL)~\cite{li2020generalized} simplifies the underlying continuous distribution of box locations as a discretized probability distribution. It considers ambiguity and uncertainty in data without introducing any other strong priors, which is helpful to improve the box localization accuracy especially when the boundaries of the ground-truth boxes are blurred. Upon DFL, DFLv2~\cite{li2021generalized} develops a lightweight sub-network to leverage the close correlation between distribution statistics and the real localization quality, which further boosts the detection performance. However, DFL usually outputs 17${\times}$ more regression values than general box regression, leading to a substantial overhead. The extra computation cost significantly hinders the training of small models. Whilst DFLv2 further increases the computation burden because of the extra sub-network. In our experiments, DFLv2 brings similar performance gain to DFL on our models. Consequently, we only adopt DFL in YOLOv6-M/L. Experimental details can be found in~\cref{sec:exp:ablate:loss}.
  
  \subsubsection{Object Loss}
  Object loss was first proposed in FCOS~\cite{tian2019fcos} to reduce the score of low-quality bounding boxes so that they can be filtered out in post-processing. 
  % Specifically, foreground locations are predicted by adding a single-layer branch in parallel with the classification branch. During the training phase, the model is optimized by computing the cross-entropy loss. In the inference stage, the predictions of the Object branch are multiplied by the predictions of the corresponding classification branches, which are output as the final classification score. 
  It was also used in YOLOX~\cite{ge2021yolox} to accelerate convergence and improve network accuracy. As an anchor-free framework like FCOS and YOLOX, we have tried object loss into YOLOv6. Unfortunately, it doesn't bring many positive effects. Details are given in~\cref{sec:exp}.

  \subsection{Industry-handy improvements}
  \label{sec:method:further}
  The following tricks come ready to use in real practice. They are not intended for a fair comparison but steadily produce performance gain without much tedious effort.
   
  \subsubsection{More training epochs}
  Empirical results have shown that detectors have a progressing performance with more training time. We extended the training duration from 300 epochs to 400 epochs to reach a better convergence.

  \subsubsection{Self-distillation}
  To further improve the model accuracy while not introducing much additional computation cost, we apply the classical knowledge distillation technique minimizing the KL-divergence between the prediction of the teacher and the student. We limit the teacher to be the student itself but pretrained, hence we call it self-distillation. Note that the KL-divergence is generally utilized to measure the difference between data distributions. However, there are two sub-tasks in object detection, in which only the classification task can directly utilize knowledge distillation based on KL-divergence. Thanks to DFL loss~\cite{li2020generalized}, we can perform it on box regression as well. The knowledge distillation loss can then be formulated as: 
 \begin{equation}
 L_{KD} = KL(p_t^{cls}||p_s^{cls}) + KL(p_t^{reg}||p_s^{reg}),
 \end{equation}
 where $p_t^{cls}$ and $p_s^{cls}$ are class prediction of the teacher model and the student model respectively, and accordingly $p_t^{reg}$ and $p_s^{reg}$ are box regression predictions. The overall loss function is now formulated as:
 \begin{equation}
 L_{total} = L_{det} + \alpha L_{KD},
 \end{equation}
 where $L_{det}$ is the detection loss computed with predictions and labels. The hyperparameter $\alpha$ is introduced to balance two losses. In the early stage of training, the soft labels from the teacher are easier to learn. As the training continues, the performance of the student will match the teacher so that the hard labels will help students more. Upon this, we apply \emph{cosine weight decay} to $\alpha$ to dynamically adjust the information from hard labels and soft ones from the teacher. We conducted detailed experiments to verify the effect of self-distillation on YOLOv6, which will be discussed in~\cref{sec:exp}.
 
 %We find the value of $\alpha$ has negligible effects on the learning efficiency.

 \subsubsection{Gray border of images}
 \label{sec:method:further:gray}
 We notice that a half-stride gray border is put around each image when evaluating the model performance in the implementations of YOLOv5~\cite{yolov5} and YOLOv7~\cite{wang2022yolov7}. Although no useful information is added, it helps in detecting the objects near the edge of the image. This trick also applies in  YOLOv6. 
 
 However, the extra gray pixels evidently reduce the inference speed. Without the gray border, the performance of YOLOv6 deteriorates, which is also the case in~\cite{yolov5,wang2022yolov7}. We postulate that the problem is related to the gray borders padding in Mosaic augmentation~\cite{bochkovskiy2020yolov4,yolov5}. Experiments on turning mosaic augmentations off during last epochs~\cite{ge2021yolox} (aka. fade strategy) are conducted for verification. In this regard, we change the area of gray border and resize the image with gray borders directly to the target image size. Combining these two strategies, our models can maintain or even boost the performance without the degradation of inference speed.

\subsection{Quantization and Deployment}\label{sec:quant-deploy}
For industrial deployment, it has been common practice to adopt quantization to further speed up runtime without much performance compromise. Post-training quantization (PTQ) directly quantizes the model with only a small calibration set. Whereas quantization-aware training (QAT) further improves the performance with the access to the training set, which is typically used jointly with distillation. However, due to the heavy use of re-parameterization blocks in YOLOv6, previous PTQ techniques fail to produce high performance, while it is hard to incorporate QAT when it comes to matching fake quantizers during training and inference. We here demonstrate the pitfalls and our cures during deployment.

\subsubsection{Reparameterizing Optimizer}

\begin{figure}[htp]
    \centering
    \includegraphics[width=0.8\columnwidth]{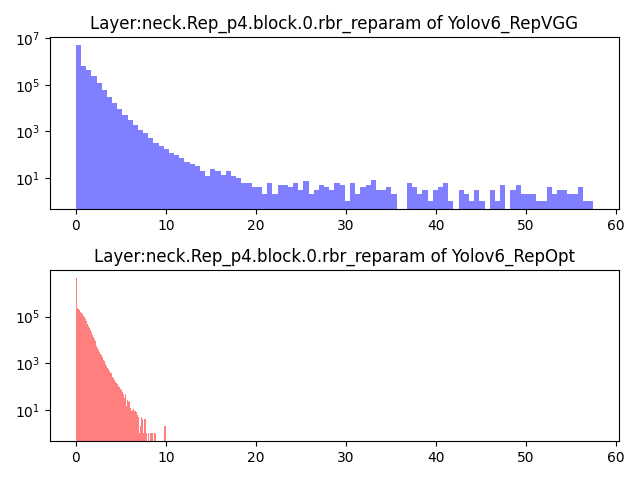}
    \caption{Improved activation distribution of YOLOv6-S trained with RepOptimizer.}
    \label{fig:repopt_act_map}
  \end{figure}

RepOptimizer \cite{ding2022re} proposes gradient re-parameterization at each optimization step.  This technique also well solves the quantization problem of reparameterization-based models. We hence reconstruct the re-parameterization blocks of YOLOv6 in this fashion and train it with RepOptimizer to obtain PTQ-friendly weights. The distribution of feature map is largely narrowed (e.g. ~\cref{fig:repopt_act_map}, more in \ref{app:quant-reopt-dist}), which greatly benefits the quantization process, see Sec~\ref{sec:sub-ptq} for results.

\subsubsection{Sensitivity Analysis}\label{quant:sens}
We further improve the PTQ performance by partially converting quantization-sensitive operations into float computation. To obtain the sensitivity distribution, several metrics are commonly used, mean-square error (MSE), signal-noise ratio (SNR) and cosine similarity. Typically for comparison, one can pick the output feature map (after the activation of a certain layer) to calculate these metrics with and without quantization. As an alternative, it is also viable to compute validation AP by switching quantization on and off for the certain layer \cite{nvidia2021pq}.

We compute all these metrics on the YOLOv6-S model trained with RepOptimizer and pick the top-6 sensitive layers to run in float. The full chart of sensitivity analysis can be found in ~\ref{app:quant-sens}.

\subsubsection{Quantization-aware Training with Channel-wise Distillation}
\begin{figure}[htp]
    \centering
    \includegraphics[width=8cm]{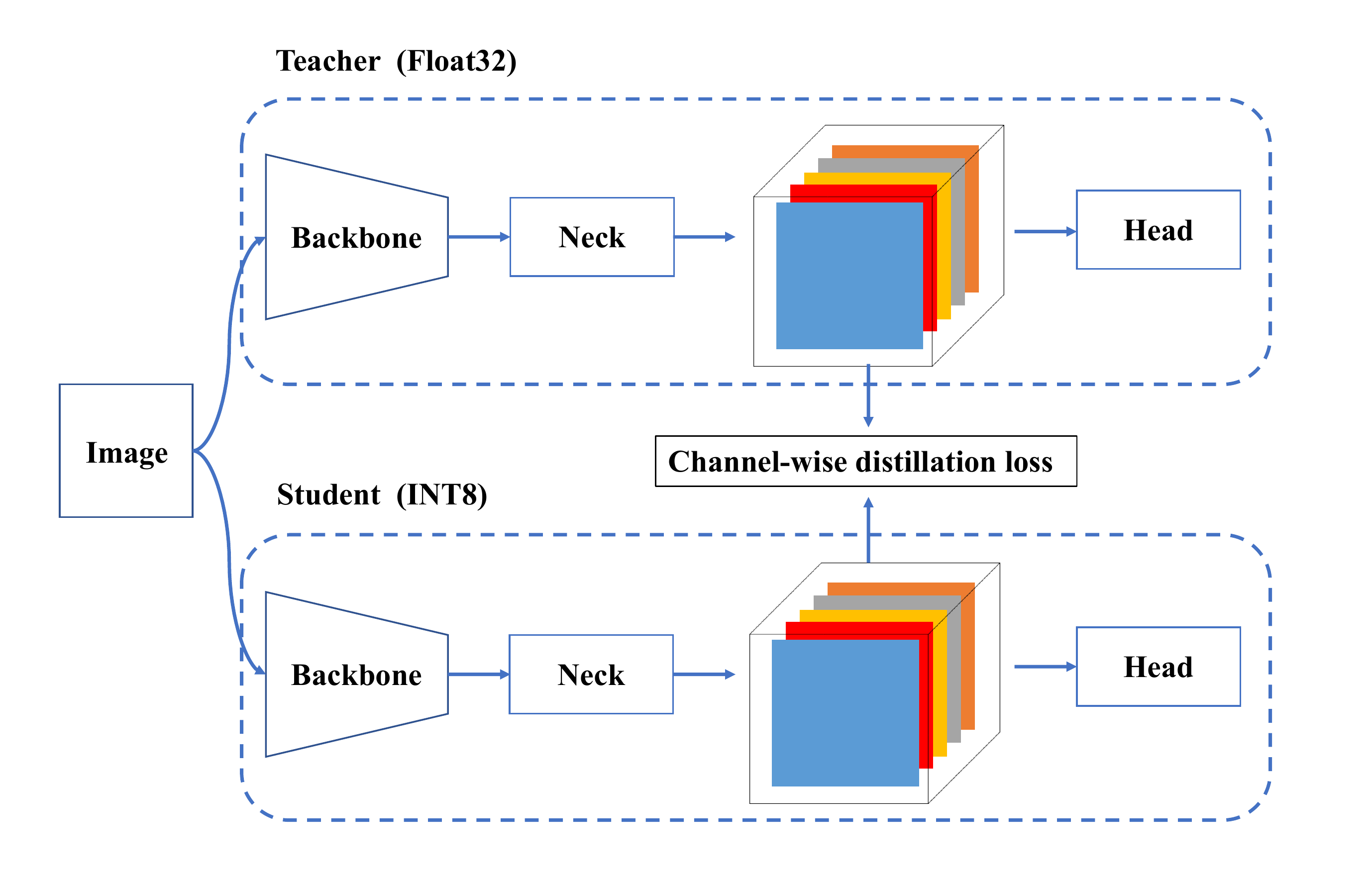}
    \caption{Schematic of YOLOv6 channel-wise distillation in QAT.}
    \label{fig:cw-distill}
  \end{figure}

In case PTQ is insufficient, we propose to involve quantization-aware training (QAT) to boost quantization performance. To resolve the problem of the inconsistency of fake quantizers during training and inference, it is necessary to build QAT upon the RepOptimizer. Besides, channel-wise distillation \cite{shu2021channel} (later as CW Distill) is adapted within the YOLOv6 framework, shown in ~\cref{fig:cw-distill}. This is also a self-distillation approach where the teacher network is the student itself in FP32-precision. See experiments in Sec~\ref{sec:sub-ptq}.

  \begin{table*}[h]
    \centering
    \resizebox{0.8\textwidth}{!}{
      \begin{tabular}{l|c|c|c|c|c|c|c|c}
        \toprule
        \multirow{2}{*}{\textbf{Method}} & \multirow{2}{*}{\textbf{Input Size}} & \multirow{2}{*}{\textbf{AP}$^{val}$} & \multirow{2}{*}{\textbf{AP}$_{50}^{val}$} &\multirow{2}{*}{\textbf{FPS}} &\multirow{2}{*}{\textbf{FPS}} &\multirow{2}{*}{\textbf{Latency}} & \multirow{2}{*}{\textbf{Params}} & \multirow{2}{*}{\textbf{FLOPs}} \\
        & & & & \tiny{(bs=1)} & \tiny{(bs=32)} &\tiny{(bs=1)} & & \\			
        \midrule
        \midrule
        YOLOv5-N~\cite{yolov5} & 640 & 28.0\% & 45.7\% &602  & 735 & 1.7 ms& 1.9 M & 4.5 G\\
        YOLOv5-S~\cite{yolov5} & 640 & 37.4\% & 56.8\% & 376 & 444 & 2.7 ms& 7.2 M & 16.5 G \\
        YOLOv5-M~\cite{yolov5} & 640 & 45.4\% & 64.1\% & 182 & 209 & 5.5 ms& 21.2 M & 49.0 G  \\
        YOLOv5-L~\cite{yolov5} & 640 & 49.0\% & 67.3\% &113  & 126 & 8.8 ms& 46.5 M & 109.1 G  \\
        \midrule
        \midrule
        YOLOX-Tiny~\cite{ge2021yolox} & 416 & 32.8\% & 50.3\%$^*$  &717  & 1143 &1.4 ms & 5.1 M & 6.5 G \\
        YOLOX-S~\cite{ge2021yolox} & 640 & 40.5\% & 59.3\%$^*$ &333  & 396 & 3.0 ms & 9.0 M & 26.8 G \\
        YOLOX-M~\cite{ge2021yolox} & 640 & 46.9\% & 65.6\%$^*$ &155  & 179 & 6.4 ms &25.3 M & 73.8 G \\
        YOLOX-L~\cite{ge2021yolox} & 640 & 49.7\% & 68.0\%$^*$ &94  & 103 & 10.6 ms &54.2 M &155.6 G \\
        \midrule
        \midrule
        PPYOLOE-S~\cite{xu2022ppyoloe} & 640 & 43.1\% & 59.6\% & 327 & 419 & 3.1 ms& 7.9 M & 17.4 G \\
        PPYOLOE-M~\cite{xu2022ppyoloe} & 640 & 49.0\% & 65.9\% & 152 & 189 & 6.6 ms& 23.4 M& 49.9 G \\
        PPYOLOE-L~\cite{xu2022ppyoloe} & 640 & 51.4\% & 68.6\% & 101 & 127 & 10.1 ms& 52.2 M& 110.1 G \\
        \midrule
        \midrule
        YOLOv7-Tiny~\cite{wang2022yolov7} & 416 &33.3\%$^*$ & 49.9\%$^*$ & 787  &1196  & 1.3 ms & 6.2 M & 5.8 G \\
        YOLOv7-Tiny~\cite{wang2022yolov7} & 640 &37.4\%$^*$ & 55.2\%$^*$ & 424  & 519 & 2.4 ms & 6.2 M & 13.7 G$^*$ \\
        YOLOv7~\cite{wang2022yolov7}  & 640 & 51.2\% & 69.7\% & 110 & 122 & 9.0 ms & 36.9 M& 104.7 G \\
        \midrule
        \midrule
        YOLOv6-N & 640 & 35.9\% & 51.2\% & 802 & 1234 &1.2 ms &4.3 M & 11.1 G \\
        YOLOv6-T & 640 & 40.3\% & 56.6\% & 449 & 659 &2.2 ms & 15.0 M & 36.7 G \\
        YOLOv6-S & 640 & 43.5\% & 60.4\% & 358 & 495 &2.8 ms & 17.2 M & 44.2 G \\
        YOLOv6-M$^\ddagger$ & 640 & 49.5\% & 66.8\%& 179 & 233 & 5.6 ms & 34.3 M & 82.2 G \\
        YOLOv6-L-ReLU$^\ddagger$ & 640 & 51.7\% & 69.2\%&  113& 149 & 8.8 ms & 58.5 M & 144.0 G \\
        YOLOv6-L$^\ddagger$ & 640 & 52.5\% & 70.0\% & 98  & 121 &10.2 ms & 58.5 M & 144.0 G \\
        \bottomrule
      \end{tabular}
    }
    \caption{
      Comparisons with other YOLO-series detectors on COCO 2017 \emph{val}. FPS and latency are measured  in FP16-precision on a Tesla T4 in the same environment with TensorRT. All our models are trained for 300 epochs without pre-training or any external data. Both the accuracy and the speed performance of our models are evaluated with the input resolution of 640$\times$640. `$\ddagger$' represents that the proposed self-distillation method is utilized. `$*$' represents the re-evaluated result of the released model through the official code.
    }
    \label{tab:sota}
  \end{table*}

\section{Experiments}
\label{sec:exp}
\subsection{Implementation Details}
We use the same optimizer and the learning schedule as YOLOv5~\cite{yolov5}, \ie stochastic gradient descent (SGD) with momentum and cosine decay on learning rate. Warm-up, grouped weight decay strategy and the exponential moving average (EMA) are also utilized. We adopt two strong data augmentations (Mosaic~\cite{bochkovskiy2020yolov4, yolov5} and Mixup~\cite{zhang2017mixup}) following~\cite{bochkovskiy2020yolov4,yolov5,ge2021yolox}. A complete list of hyperparameter settings can be found in our released code. We train our models on the COCO 2017~\cite{lin2014microsoft} training set, and the accuracy is evaluated on the COCO 2017 validation set. All our models are trained on 8 NVIDIA A100 GPUs, and the speed performance is measured on an NVIDIA Tesla T4 GPU with TensorRT version 7.2 unless otherwise stated. And the speed performance measured with other TensorRT versions or on other devices is demonstrated in Appendix \ref{app:bench}.

\subsection{Comparisons}
  Considering that the goal of this work is to build networks for industrial applications, we primarily focus on the speed performance of all models after deployment, including throughput (FPS at a batch size of 1 or 32) and the GPU latency, rather than FLOPs or the number of parameters. We compare YOLOv6 with other state-of-the-art detectors of YOLO series, including YOLOv5~\cite{yolov5}, YOLOX~\cite{ge2021yolox}, PPYOLOE~\cite{xu2022ppyoloe} and YOLOv7~\cite{wang2022yolov7}. Note that we test the speed performance of all official models with FP16-precision on the same Tesla T4 GPU with TensorRT~\cite{tensorrt}. The performance of YOLOv7-Tiny is re-evaluated according to their open-sourced code and weights at the input size of 416 and 640. Results are shown in~\cref{tab:sota} and~\cref{fig:sota-comp}. Compared with YOLOv5-N/YOLOv7-Tiny (input size=416), our YOLOv6-N has significantly advanced by 7.9\%/2.6\% respectively. It also comes with the best speed performance in terms of both throughput and latency. Compared with YOLOX-S/PPYOLOE-S, YOLOv6-S can improve AP by 3.0\%/0.4\% with higher speed. We compare YOLOv5-S and YOLOv7-Tiny (input size=640) with YOLOv6-T, our method is 2.9\% more accurate and 73/25 FPS faster with a batch size of 1. YOLOv6-M outperforms YOLOv5-M by 4.2\% higher AP with a similar speed, and it achieves 2.7\%/0.6\% higher AP than YOLOX-M/PPYOLOE-M at a higher speed. Besides, it is more accurate and faster than YOLOv5-L.  YOLOv6-L is 2.8\%/1.1\% more accurate than YOLOX-L/PPYOLOE-L under the same latency constraint. We additionally provide a faster version of YOLOv6-L by replacing SiLU with ReLU (denoted as YOLOv6-L-ReLU). It achieves 51.7\% AP with a latency of 8.8 ms, outperforming YOLOX-L/PPYOLOE-L/YOLOv7 in both accuracy and speed.

\subsection{Ablation Study}
\subsubsection{Network}
\paragraph{Backbone and neck}
  We explore the influence of single-path structure and multi-branch structure on backbones and necks, as well as the channel coefficient (denoted as CC) of CSPStackRep Block. All models described in this part adopt TAL as the label assignment strategy, VFL as the classification loss, and GIoU with DFL as the regression loss. Results are shown in~\cref{tab:ablate:network}. We find that the optimal network structure for models at different sizes should come up with different solutions. 

  For YOLOv6-N, the single-path structure outperforms the multi-branch structure in terms of both accuracy and speed. Although the single-path structure has more FLOPs and parameters than the multi-branch structure, it could run faster due to a relatively lower memory footprint and a higher degree of parallelism. For YOLOv6-S, the two block styles bring similar performance. When it comes to larger models, multi-branch structure achieves better performance in accuracy and speed. And we finally select multi-branch with a channel coefficient of 2/3 for YOLOv6-M and 1/2 for YOLOv6-L. 

  Furthermore, we study the influence of width and depth of the neck on YOLOv6-L. Results in~\cref{tab:ablate:neck} show that the slender neck performs 0.2\% better than the wide-shallow neck with the similar speed.

  \begin{table}
      \centering
      \scalebox{0.7}{
            \begin{tabular}{l|c|c|c|c|c|c}
              \toprule
              \multirow{2}{*}{\textbf{Models}}   & \multirow{2}{*}{\textbf{Block}} & \multirow{2}{*}{\textbf{CC}} & \multirow{2}{*}{\textbf{AP$^{val}$}} & \multirow{2}{*}{\textbf{FPS}} & \multirow{2}{*}{\textbf{Params}} & \multirow{2}{*}{\textbf{FLOPs}} \\
              & & &  &  \tiny{(bs=32)} & & \\
              \midrule
              \midrule
              \multirow{2}{*}{YOLOv6-N}
              &RepBlock & - & 35.2\% & 1237 & 4.3M& 11.1G \\
              &CSPStackRep Block & 1/2 & 32.7\% & 1257 & 2.3M& 5.6G \\
              \midrule
              \multirow{2}{*}{YOLOv6-S}
              &RepBlock & - &43.2\% &499 & 17.2M & 44.2G \\
              &CSPStackRep Block & 1/2 &43.4\% &511 & 11.5M & 27.7G \\
              \midrule
              \multirow{3}{*}{YOLOv6-M}
              &RepBlock & - &47.9\% &137 & 67.1M & 175.6G \\
              &CSPStackRep Block & 2/3 &48.1\% &237 & 34.3M & 82.2G \\
              &CSPStackRep Block & 1/2 &47.3\% &237 & 27.7M & 68.4G \\
              \midrule
              \multirow{2}{*}{YOLOv6-L}
              &CSPStackRep Block & 2/3 &50.1\% &149 & 54.7M & 142.7G \\
              &CSPStackRep Block & 1/2 &50.1\% &151 & 58.5M & 144.0G \\
            \bottomrule
            \end{tabular}
      }
    \caption{Ablation study on backbones and necks. YOLOv6-L here is equipped with  ReLU.}
    \label{tab:ablate:network}
\end{table}

  \begin{table}[!thbp]
    \begin{center}
      \scalebox{0.45}{
        \resizebox{\textwidth}{!}{
            \begin{tabular}{cccccc}
              \toprule
              \multirow{2}{*}{\textbf{Width}} & \multirow{2}{*}{\textbf{Depth}} & \multirow{2}{*}{\textbf{AP$^{val}$}} & \multirow{2}{*}{\textbf{FPS}} & \multirow{2}{*}{\textbf{Params}} & \multirow{2}{*}{\textbf{FLOPs}} \\
              & & &  \tiny{(bs=32)} & & \\
              \midrule
              \midrule
                    $[192, 384, 768]$ & 5 & 50.8\% & 123 & 58.6 M & 144.7 G \\
                    $[128, 256, 512]$ & 12 & 51.0\% & 122 & 58.5 M & 144.0 G \\
            \bottomrule
            \end{tabular}
      }}
    \end{center}
    \caption{Ablation study on the neck settings of YOLOv6-L. SiLU is selected as the activation function.}
    \label{tab:ablate:neck}
  \end{table}

  \paragraph{Combinations of convolutional layers and activation functions}
 YOLO series adopted a wide range of activation functions, ReLU~\cite{nair2010rectified}, LReLU~\cite{maas2013rectifier}, Swish~\cite{ramachandran2017searching}, SiLU~\cite{elfwing2018sigmoid}, Mish~\cite{misra2019mish} and so on. Among these activation functions, SiLU is the most used. Generally speaking, SiLU performs with better accuracy and does not cause too much extra computation cost. However, when it comes to industrial applications, especially for deploying models with TensorRT~\cite{tensorrt} acceleration, ReLU has a greater speed advantage because of its fusion into convolution. 
 Moreover, we further verify the effectiveness of combinations of RepConv/ordinary convolution (denoted as Conv) and ReLU/SiLU/LReLU in networks of different sizes to achieve a better trade-off. As shown in~\cref{tab:ablate:actconv}, Conv with SiLU performs the best in accuracy while the combination of RepConv and ReLU achieves a better trade-off. We suggest users adopt RepConv with ReLU in latency-sensitive applications. We choose to use RepConv/ReLU combination in YOLOv6-N/T/S/M for higher inference speed and use the Conv/SiLU combination in the large model YOLOv6-L to speed up training and improve performance.

  \begin{table}
    \centering
    \scalebox{0.9}{
          \begin{tabular}{l|c|c|c|c}
            \toprule
            \multirow{2}{*}{\textbf{Model}} &  \multirow{2}{*}{\textbf{Conv.}} &  \multirow{2}{*}{\textbf{Act.}} &  \multirow{2}{*}{\textbf{AP$^{val}$}} &  \multirow{2}{*}{\textbf{FPS}} \\
            & & & & \tiny{(bs=32)} \\
            \midrule
            \midrule
            \multirow{6}{*}{YOLOv6-N} &Conv & SiLU & \textbf{36.6}\% & 963  \\
                                      &RepConv & SiLU & 36.5\%  & 971 \\
                                      &Conv & ReLU & 34.8\% &\textbf{1246}  \\
                                      &RepConv & ReLU & 35.2\% & 1233 \\
                                      &Conv & LReLU &  35.4\% &983 \\
                                      &RepConv & LReLU &35.6\% & 975 \\
            \midrule
            \multirow{6}{*}{YOLOv6-M} &Conv & SiLU & \textbf{48.9}\% &180  \\
                                      &RepConv & SiLU & \textbf{48.9}\% &180 \\
                                      &Conv & ReLU & 47.7\% &235 \\
                                      &RepConv & ReLU & 48.1\% &\textbf{236}\\
                                      &Conv & LReLU & 48.0\% &185 \\
                                      &RepConv & LReLU &48.1\% & 187 \\
            
          \bottomrule
          \end{tabular}
    }
  \caption{Ablation study on combinations of different types of convolutional layers (denoted as Conv.) and activation layers (denoted as Act.).}
  \label{tab:ablate:actconv}
\end{table}

  \paragraph{Miscellaneous design}
  We also conduct a series of ablation on other network parts mentioned in~\cref{sec:method:network} based on YOLOv6-N. We choose YOLOv5-N as the baseline and add other components incrementally. Results are shown in~\cref{tab:ablate:network_other}. Firstly, with decoupled head (denoted as DH), our model is 1.4\% more accurate with 5\% increase in time cost. Secondly, we verify that the anchor-free paradigm is 51\% faster than the anchor-based one for its 3$\times$ less predefined anchors, which results in less dimensionality of the output. Further, the unified modification of the backbone (EfficientRep Backbone) and the neck (Rep-PAN neck), denoted as EB+RN, brings 3.6\% AP improvements, and runs 21\% faster. Finally, the optimized decoupled head (hybrid channels, HC) brings 0.2\% AP and 6.8\% FPS improvements in accuracy and speed respectively.

  \begin{table}
    \centering
    \scalebox{0.9}{
          \begin{tabular}{l|c|c|c|c|c}
            \toprule
            \multirow{2}{*}{\textbf{DH}} & \multirow{2}{*}{\textbf{AF}} & \multirow{2}{*}{\textbf{EB+RN}} & \multirow{2}{*}{\textbf{HC}} & \multirow{2}{*}{\textbf{AP$^{val}$}} & \multirow{2}{*}{\textbf{FPS}} \\
            &&&&&\tiny{(bs=32)} \\
            \midrule
            \midrule
            \XSolidBrush&\XSolidBrush &\XSolidBrush &\XSolidBrush &  28.0\%& 672\\
            \Checkmark &\XSolidBrush &\XSolidBrush &\XSolidBrush &  29.4\% &637 \\
            \Checkmark & \Checkmark &\XSolidBrush &\XSolidBrush &  30.7\%&962 \\
            \Checkmark & \Checkmark & \Checkmark &\XSolidBrush &  34.3\%&1163 \\
            \Checkmark & \Checkmark & \Checkmark & \Checkmark &  \textbf{34.5}\%&\textbf{1242} \\
            % \Checkmark & \Checkmark & \Checkmark & \Checkmark & \Checkmark & 35.1&1143 \\
          \bottomrule
          \end{tabular}
    }
    \caption{Ablation study on all network designs in an incremental way. FPS is tested with FP16-precision and batch-size=32 on Tesla T4 GPUs.}
    \label{tab:ablate:network_other}
  \end{table}

\subsubsection{Label Assignment}
\label{sec:exp:ablate:labelassign}
In~\cref{tab:label_assign}, we analyze the effectiveness of mainstream label assign strategies. Experiments are conducted on YOLOv6-N. As expected, we observe that SimOTA and TAL are the best two strategies. Compared with the ATSS, SimOTA can increase AP by 2.0\%, and TAL brings 0.5\% higher AP than SimOTA. Considering the stable training and better accuracy performance of TAL, we adopt TAL as our label assignment strategy.

 \begin{table}
  \centering
  \begin{tabular}{l|c}
    \toprule
    \textbf{Method} & \textbf{AP$^{val}$} \\
    \midrule
    \midrule
    ATSS~\cite{zhang2020atss} & 32.5\%  \\
    SimOTA~\cite{ge2021yolox} & 34.5\% \\
    TAL~\cite{feng2021tood} & \textbf{35.0}\% \\
    DW~\cite{dw_CVPR} & 33.4\%  \\
    ObjectBox~\cite{Zand2022ObjectBoxFC} & 30.1\% \\
    \bottomrule
  \end{tabular}
  \caption{Comparisons of label assignment methods.}
  \label{tab:label_assign}
\end{table}

% We carried out other ablation experiments on TAL, including variants of the head structure and warm-up method. For ESE head, the experiments shows that it slows down the inference speed by 15\% and 19\% under batchsize=1 and batchsize=32, respectively. 
% Therefore, we abandon modification of the head structure.

In addition, the implementation of TOOD~\cite{feng2021tood} adopts ATSS~\cite{zhang2020atss} as the warm-up label assignment strategy during the early training epochs. We also retain the warm-up strategy and further make some explorations on it. Details are shown in~\cref{tab:warmup}, and we can find that without warm-up or warmed up by other strategies (i.e., SimOTA) it can also achieve the similar performance.

\begin{table}
  \centering
  \scalebox{0.9}{
    \begin{tabular}{cc}
      \toprule
      \textbf{Warmup strategy} &\textbf{AP$^{val}$}  \\
      \midrule
      \midrule
      w/o& 34.9\%  \\
      ATSS~\cite{zhang2020atss}& \textbf{35.0}\%  \\
      SimOTA~\cite{ge2021yolox}& 34.9\% \\
      \bottomrule
    \end{tabular}
  }
  \caption{Comparisons of label assignment methods in warm-up stage.}
  \label{tab:warmup}
\end{table}

% \subsubsection{DW}
% In order to find a more suitable label matching strategy for YOLOv6, we also try to introduce the DW strategy to improve the effect of the model. From the experimental results, when DW is paired with YOLOv6, the effect is not ideal. For example, on the n network, it is down about 1.1 mAP compared to SimOTA and about 1.6 mAP compared to TAL. The possible reason for this negative return is that...

% \subsubsection{Objectbox}
% We also tried the objectbox of ECCV2022. However, it does not show competitiveness compared to other methods. In fact, mAP is poor (about -4\%). The reason for this may be that the location-based static label assignment ignores the dynamically changing features of anchors, resulting in suboptimal selection of positive samples.

\subsubsection{Loss functions}
\label{sec:exp:ablate:loss}
In the object detection framework, the loss function is composed of a classification loss, a box regression loss and an optional object loss, which can be formulated as follows:
\begin{equation}
L_{det} = L_{cls} + \lambda L_{reg} + \mu L_{obj},
\end{equation}
where $L_{cls}$, $L_{reg}$ and $L_{obj}$ are classification loss, regression loss and object loss. $\lambda$ and $\mu$ are hyperparameters. 

In this subsection, we evaluate each loss function on YOLOv6. Unless otherwise specified, the baselines for YOLOv6-N, YOLOv6-S and YOLOv6-M are 35.0\%, 42.9\% and 48.0\% trained with TAL, Focal Loss and GIoU Loss.
%  \begin{table}
%   \centering
%   \begin{tabular}{@{}lccc@{}}
%     \toprule
%     Loss/Model & YOLOv6-N & YOLOv6-S & YOLOv6-M  \\
%     \midrule
%     Focal loss & 35.0\% & 42.9\% & 48.0\%  \\
%     Poly loss & - & - & - \\
%     Quality Focal loss & - & - & -\\
%     VariFocal loss & 35.2\% & 43.2\% & 48.1\% \\
%     \bottomrule
%   \end{tabular}
%   \caption{Classification loss comparison(300 epoch).}
%   \label{tab:ablate:loss}
% \end{table}

\begin{table}
  \centering
  \scalebox{0.9}{
        \begin{tabular}{l|c|c}
          \toprule
          \textbf{Model}   & \textbf{Classification Loss} & \textbf{AP$^{val}$} \\
          \midrule
          \midrule
          \multirow{4}{*}{YOLOv6-N} & Focal Loss~\cite{lin2017focal} &35.0\%  \\
           & Poly Loss~\cite{leng2022polyloss} & 34.0\%   \\
           & QFL~\cite{li2020generalized} & \textbf{35.4}\%  \\
           & VFL~\cite{zhang2021varifocalnet} &  35.2\% \\
          \midrule
          \multirow{4}{*}{YOLOv6-S} & Focal Loss~\cite{lin2017focal} & 42.9\%  \\
           & Poly Loss~\cite{leng2022polyloss} &  41.5\% \\
           & QFL~\cite{li2020generalized} & 	43.1\%  \\
           & VFL~\cite{zhang2021varifocalnet} &  \textbf{43.2}\% \\
           \midrule
           \multirow{4}{*}{YOLOv6-M} & Focal Loss~\cite{lin2017focal} &  48.0\% \\
            & Poly Loss~\cite{leng2022polyloss} &  46.9\% \\
            & QFL~\cite{li2020generalized} &  48.0\% \\
            & VFL~\cite{zhang2021varifocalnet} &  \textbf{48.1}\% \\

        \bottomrule
        \end{tabular}
  }
\caption{Ablation study on classification loss functions.}
\label{tab:ablate:clsloss}
\end{table}

\paragraph{Classification Loss} We experiment Focal Loss~\cite{lin2017focal}, Poly loss\cite{leng2022polyloss}, QFL~\cite{li2020generalized} and VFL~\cite{zhang2021varifocalnet} on YOLOv6-N/S/M. As can be seen in~\cref{tab:ablate:clsloss}, VFL brings 0.2\%/0.3\%/0.1\% AP improvements on YOLOv6-N/S/M respectively compared with Focal Loss. We choose VFL as the classification loss function.

\paragraph{Regression Loss} IoU-series and probability loss functions are both experimented with on YOLOv6-N/S/M.

The latest IoU-series losses are utilized in YOLOv6-N/S/M. Experiment results in~\cref{tab:ablate:iouloss} show that SIoU Loss outperforms others for YOLOv6-N and YOLOv6-T, while CIoU Loss performs better on YOLOv6-M.

For probability losses, as listed in~\cref{tab:ablate:probloss}, introducing DFL can obtain 0.2\%/0.1\%/0.2\% performance gain for YOLOv6-N/S/M respectively. However, the inference speed is greatly affected for small models. Therefore, DFL is only introduced in YOLOv6-M/L.

\begin{table}[ht]
  \centering
  \scalebox{0.9}{
        \begin{tabular}{l|c|c}
          \toprule
          \textbf{Model} & \textbf{Loss} & \textbf{AP$^{val}$} \\
          \midrule
          \midrule
          \multirow{3}{*}{YOLOv6-N} & GIoU~\cite{rezatofighi2019generalized} & 35.1\%  \\
           & CIoU~\cite{zheng2020distance} &  35.1\%  \\
           & SIoU~\cite{gevorgyan2022siou} &  \textbf{35.5}\% \\
          \midrule
          \multirow{3}{*}{YOLOv6-S} & GIoU~\cite{rezatofighi2019generalized} & 43.1\% \\
           & CIoU~\cite{zheng2020distance} &  43.1\%  \\
           & SIoU~\cite{gevorgyan2022siou} &  \textbf{43.3}\% \\ 
           \midrule
          \multirow{3}{*}{YOLOv6-M} & GIoU~\cite{rezatofighi2019generalized} & 48.2\% \\
           & CIoU~\cite{zheng2020distance} &  \textbf{48.3}\%  \\
           & SIoU~\cite{gevorgyan2022siou} &  48.1\% \\ 
        \bottomrule
        \end{tabular}
  }
\caption{Ablation study on IoU-series box regression loss functions. The classification loss is VFL~\cite{zhang2021varifocalnet}.}
\label{tab:ablate:iouloss}
\end{table}

\begin{table}[ht]
  \centering
  \scalebox{0.9}{
    \begin{tabular}{l|cccc}
      \toprule
      \multirow{2}{*}{\textbf{Method}} & \multirow{2}{*}{\textbf{Loss}} & \multirow{2}{*}{\textbf{AP$^{val}$}} & \multirow{2}{*}{\textbf{FPS}} \\
        & & &  \tiny{(bs=32)} \\
      \midrule
      \midrule
      \multirow{3}{*}{YOLOv6-N} & w/o & 35.0\% &  \textbf{1226}  \\
        & DFL~\cite{li2020generalized} & \textbf{35.2}\% &  1022  \\
        & DFLv2~\cite{li2021generalized} & \textbf{35.2}\%  &819  \\
      \midrule
      \multirow{3}{*}{YOLOv6-S} & w/o & 42.9\% &  \textbf{486}  \\
        & DFL~\cite{li2020generalized} & \textbf{43.0}\%  & 461  \\
        & DFLv2~\cite{li2021generalized} & \textbf{43.0}\% & 422 \\
      \midrule
      \multirow{3}{*}{YOLOv6-M} & w/o & 48.0\% & 233 \\
        & DFL~\cite{li2020generalized} & 48.2\% &  \textbf{236}  \\
        & DFLv2~\cite{li2021generalized} & \textbf{48.3}\% & 226 \\
      \bottomrule
    \end{tabular}
  }
\caption{Ablation study on probability loss functions.}
\label{tab:ablate:probloss}
\end{table}

\paragraph{Object Loss}
Object loss is also experimented with YOLOv6, as shown in~\cref{tab:ablate:objloss}. From~\cref{tab:ablate:objloss}, we can see that object loss has negative effects on YOLOv6-N/S/M networks, where the maximum decrease is 1.1\% AP on YOLOv6-N. The negative gain may come from the conflict between the object branch and the other two branches in TAL. Specifically, in the training stage, IoU between predicted boxes and ground-truth ones, as well as classification scores are used to jointly build a metric as the criteria to assign labels. However, the introduced object branch extends the number of tasks to be aligned from two to three, which obviously increases the difficulty. Based on the experimental results and this analysis, the object loss is then discarded in YOLOv6.

 \begin{table}
  \centering
  \scalebox{0.9}{
    \begin{tabular}{l|c|c}
      \toprule
      \textbf{Method} & \textbf{Object Loss} & \textbf{AP$^{val}$} \\
      \midrule
      \midrule
      \multirow{2}{*}{YOLOv6-N} & \XSolidBrush & \textbf{35.0}\%  \\
      & \Checkmark &  33.9\% \\
      \midrule
      \multirow{2}{*}{YOLOv6-S} & \XSolidBrush& \textbf{42.9}\% \\
      & \Checkmark & 41.4\% \\
      \midrule
      \multirow{2}{*}{YOLOv6-M} & \XSolidBrush & \textbf{48.0}\% \\
      & \Checkmark & 46.5\% \\
      \bottomrule
    \end{tabular}
  }
  \caption{Effectiveness of object loss.}
  \label{tab:ablate:objloss}
\end{table}

% \subsection{Data augmentations}
% We discuss the effectiveness of \emph{Mosaic augmentation} and \emph{Mixup augmentation} in this section. 

% \subsubsection{Mosaic Augmentation}
% \label{subsec:mosaic}
% The experiments about Mosaic augmentation focus on finding the appropriate probabilities of opening Mosaic for models of different sizes, as well as the effectiveness of closing it during the last training epochs. Results of Mosaic probabilities are shown in \textcolor{red}{Fig. X}, from which we observe that networks of different sizes should open Mosaic with different probabilities. Experimental results on the strategy of closing Mosaic during last epochs~\cite{ge2021yolox} are shown in Tab.~\cref{tab:stopmosaic}, from which we know that \textcolor{red}{XXX}.

% In addition, the effectiveness of the strategy for enhancing the ability of detecting objects near the edge mentioned in~\cref{sec:method:aug} is evaluated. The experimental results are shown in \textcolor{red}{Table XX}

% \subsubsection{Mixup Augmentation}
% In experiments on Mixup, we concern with the influence of extra augmentations added to two original images before mixing. Besides, we find the distribution of interpolation coefficient also important. The results are shown in~\textcolor{red}{Fig.X}. It can be found that the \textcolor{red}{combination X} is the best way to apply augmentations to two images. 

\subsection{Industry-handy improvements}
\label{sec:ablate:further}
%The experiment results of other training strategies and tricks for further improvements are discussed in this subsection.

\begin{table}[ht]
\centering
\scalebox{0.9}{
\begin{tabular}{l|cc}
  \toprule
  \textbf{Model} & \textbf{300 epochs} & \textbf{400 epochs}\\
  \midrule
  \midrule
  YOLOv6-N & 35.9\% & 36.3\% \\
  YOLOv6-T & 40.3\% & 40.9\% \\
  YOLOv6-S & 43.4\% & 43.9\% \\
  \bottomrule
\end{tabular}
}
\caption{Experiments of more training epochs on small models.}
\label{tab:moreepoch}
\end{table}
\paragraph{More training epochs}
In practice, more training epochs is a simple and effective way to further increase the accuracy. Results of our small models trained for 300 and 400 epochs are shown in~\cref{tab:moreepoch}. We observe that training for longer epochs substantially boosts AP by 0.4\%, 0.6\%, 0.5\% for YOLOv6-N, T, S respectively. Considering the acceptable cost and the produced gain, it suggests that training for 400 epochs is a better convergence scheme for YOLOv6.

\begin{table}[ht]
\centering
\scalebox{0.9}{
  \begin{tabular}{cccc}
    \toprule
    \textbf{Cls.} & \textbf{Reg.} & \textbf{Weight Decay} & \textbf{AP$^{val}$} \\
    \midrule
    \midrule
    \XSolidBrush& \XSolidBrush&\XSolidBrush &51.0\%  \\
    \Checkmark &\XSolidBrush &\XSolidBrush &51.4\%  \\
    \Checkmark &\Checkmark &\XSolidBrush &51.7\%  \\
    \Checkmark &\Checkmark &\Checkmark &\textbf{52.3}\%  \\
    \bottomrule
  \end{tabular}
}
\caption{Ablation study on the self-distillation.}
\label{tab:distillation}
\end{table}
\paragraph{Self-distillation}
We conducted detailed experiments to verify the proposed self-distillation method on YOLOv6-L. As can be seen in~\cref{tab:distillation}, applying the self-distillation only on the classification branch can bring 0.4\% AP improvement. Furthermore, we simply perform the self-distillation on the box regression task to have 0.3\% AP increase. The introduction of weight decay boosts the model by 0.6\% AP.

\paragraph{Gray border of images}
  In~\cref{sec:method:further:gray}, we introduce a strategy to solve the problem of performance degradation without extra gray borders. Experimental results are shown in~\cref{tab:stopmosaic}. In these experiments, YOLOv6-N and YOLOv6-S are trained for 400 epochs and YOLOv6-M for 300 epochs. It can be observed that the accuracy of YOLOv6-N/S/M is lowered by 0.4\%/0.5\%/0.7\% without Mosaic fading when removing the gray border. However, the performance degradation becomes 0.2\%/0.5\%/0.5\% when adopting Mosaic fading, from which we find that, on the one hand, the problem of performance degradation is mitigated. On the other hand, the accuracy of small models (YOLOv6-N/S) is improved whether we pad gray borders or not. Moreover, we limit the input images to 634$\times$634 and add gray borders by 3 pixels wide around the edges (more results can be found in Appendix \ref{app:gray-border}). With this strategy, the size of the final images is the expected 640$\times$640. The results in~\cref{tab:stopmosaic} indicate that the final performance of YOLOv6-N/S/M is even 0.2\%/0.3\%/0.1\% more accurate with the final image size reduced from 672 to 640.

\begin{table}[ht]
  \centering
  \scalebox{1}{
  \scalebox{0.9}{
    \begin{tabular}{l|ccccc}
      \toprule
      \multirow{2}{*}{\textbf{Model}} & \textbf{Image} & \textbf{Border} &\textbf{Close} & \multirow{2}{*}{\textbf{AP$^{val}$}} \\
      & \textbf{Size} & \textbf{Size} & \textbf{Mosaic} & \\

      \midrule
      \midrule
      \multirow{5}*{YOLOv6-N} 
      & 640 & 16 &\XSolidBrush & 36.0\% \\
      & 640 & 16 & \Checkmark & \textbf{36.2}\% \\ 
      & 640 & 0 &\XSolidBrush & 35.6\% \\ 
      & 640 & 0 & \Checkmark & 36.0\% \\
      & 634 & 3 & \Checkmark & \textbf{36.2}\% \\
      \midrule
      \multirow{5}*{YOLOv6-S} 
      & 640 & 16 &\XSolidBrush & 43.4\% \\
      & 640 & 16 & \Checkmark & \textbf{43.9}\% \\ 
      & 640 & 0 &\XSolidBrush & 42.9\% \\ 
      & 640 & 0 & \Checkmark & 43.4\% \\
      & 634 & 3 & \Checkmark & 43.7\% \\
      \midrule
      \multirow{5}*{YOLOv6-M} 
      & 640 & 16 &\XSolidBrush & 48.4\% \\
      & 640 & 16 & \Checkmark & 48.4\% \\ 
      & 640 & 0 &\XSolidBrush & 47.7\% \\ 
      & 640 & 0 & \Checkmark & 47.9\% \\
      & 634 & 3 & \Checkmark & \textbf{48.5}\% \\
      \bottomrule
    \end{tabular}
  }
  }
  \caption{Experimental results about the strategies for solving the problem of the performance degradation without extra gray border.}
  \label{tab:stopmosaic}
\end{table}

% \begin{table}
%   \centering
%   \scalebox{0.9}{
%   \begin{tabular}{l|ccc}
%     \toprule
%     \multirow{2}{*}{\textbf{Model}} & \textbf{Image} & \textbf{Border}& \multirow{2}{*}{\textbf{AP$^{val}$}} \\
%     & \textbf{Size} & \textbf{Size}  & \\

%     \midrule
%     \midrule
%     \multirow{3}*{YOLOv6-N} 
%      & 640 & 32 &36.2\%  \\
%      & 640 & 0 &36.0\%  \\
%      & 634 & 6 & 36.2\%   \\ 
%      \midrule
%      \multirow{3}*{YOLOv6-S} 
%      & 640 & 32 &43.9\%  \\
%      & 640 & 0 &43.4\%  \\
%      & 634 & 6 & 43.7\%  \\ 

%      \midrule
%      \multirow{3}*{YOLOv6-M} 
%      & 640 & 32 &48.4\%  \\
%      & 640 & 0 &47.9\% \\
%      & 634 & 6 & 48.5\%  \\ 

%     \bottomrule
%   \end{tabular}
%   }
%   \caption{The effectiveness of adjusting the size of the gray border.}
%   \label{tab:grayborder}
% \end{table}

\subsection{Quantization Results}
We take YOLOv6-S as an example to validate our quantization method. The following experiment is on both two releases. The baseline model is trained for 300 epochs. 

\subsubsection{PTQ}\label{sec:sub-ptq}
The average performance is substantially improved when the model is trained with RepOptimizer, see Table~\ref{tab:ablate:repopt}. RepOptimizer is in general faster and nearly identical.
  \begin{table}[ht]
    \begin{center}
      \scalebox{0.45}{
        \resizebox{\textwidth}{!}{
            \begin{tabular}{l|cc}
              \toprule
              \textbf{Model} & \textbf{FP32 AP$^{val}$} & \textbf{PTQ INT8 AP$^{val}$} \\
              \midrule
              \midrule
               (v1.0)  \\
                YOLOv6-S   &  42.4 & 35.0 \\
                YOLOv6-S w/ RepOptimizer  &  42.4 & 40.9  \\
                \midrule
                (v2.0) \\
                YOLOv6-S   &  43.4 & 41.3  \\
                YOLOv6-S   w/ RepOptimizer  &  43.1 & 42.6  \\
            \bottomrule
            \end{tabular}
      }}
    \end{center}
    \caption{PTQ performance of YOLOv6s trained with RepOptimizer.}
    \label{tab:ablate:repopt}
  \end{table}

\subsubsection{QAT}\label{sec:sub-qat}
For v1.0, we apply fake quantizers to non-sensitive layers obtained from Section~\ref{quant:sens} to perform quantization-aware training and call it partial QAT. We compare the result with full QAT in Table~\ref{tab:ablate:qat-ablation}.  Partial QAT leads to better accuracy with a slightly reduced throughput.
  
 \begin{table}[ht]
    \begin{center}
      \scalebox{0.45}{
        \resizebox{\textwidth}{!}{
            \begin{tabular}{l|cc}
              \toprule
              \textbf{Model} & \textbf{INT8 AP$^{val}$} & \textbf{FPS} \\
              \midrule
              \midrule
                YOLOv6-S    &  35.0 & 556 \\
                YOLOv6-S + RepOpt + Partial QAT + CW Distill &  42.3 & 503  \\
                YOLOv6-S + RepOpt + QAT + CW Distill &  42.1 & 528  \\
            \bottomrule
            \end{tabular}
      }}
    \end{center}
    \caption{QAT performance of YOLOv6-S (v1.0) under different settings.}
    \label{tab:ablate:qat-ablation}
  \end{table}

Due to the removal of quantization-sensitive layers in v2.0 release, we directly use full QAT on YOLOv6-S trained with RepOptimizer. We eliminate inserted quantizers through graph optimization to obtain higher accuracy and faster speed. We compare the distillation-based quantization results from PaddleSlim\cite{baidu2022paddleslim}  in Table~\ref{tab:ablate:qat-ablation-v2}. Note our quantized version of YOLOv6-S is the fastest and the most accurate, also see~\cref{fig:sota-comp}.
 
 \begin{table}[ht]
    \begin{center}
      \scalebox{0.45}{
        \resizebox{\textwidth}{!}{
            \begin{tabular}{l|ccc}
              \toprule
              \textbf{Model} & \textbf{AP$^{val}$} & \textbf{FPS$^{bs=1}$} & \textbf{FPS$^{bs=32}$}  \\
              \midrule
               \midrule
               YOLOv5-S \cite{baidu2022paddleslim} & 36.9 & 502$^\dagger$&  N/A \\
               YOLOv6-S$^*$ \cite{baidu2022paddleslim} & 41.3 & 579$^\dagger$ & N/A \\
               YOLOv7-Tiny \cite{baidu2022paddleslim}  & 37.0 & 512$^\dagger$ & N/A \\
               \midrule
               YOLOv6-S (FP16) & 43.4 & 377$^\dagger$ &  541$^\dagger$ \\
               YOLOv6-S (Our QAT strategy) & 43.3 & 596$^\dagger$ & 869$^\dagger$  \\
            \bottomrule
            \end{tabular}
      }}
    \end{center}
    \caption{QAT performance of YOLOv6-S (v2.0) compared with other quantized detectors. `$^*$': based on v1.0 release. `$^\dagger$': We tested with TensorRT 8 on Tesla T4 with a batch size of 1 and 32.}
    \label{tab:ablate:qat-ablation-v2}
  \end{table}

\section{Conclusion}
In a nutshell, with the persistent industrial requirements in mind, we present the current form of YOLOv6, carefully examining all the advancements of components of object detectors up to date, meantime instilling our thoughts and practices. The result surpasses other available real-time detectors in both accuracy and speed. For the convenience of the industrial deployment, we also supply a customized quantization method for YOLOv6, rendering an ever-fast detector out-of-box. We sincerely thank the academic and industrial community for their brilliant ideas and endeavors. In the future, we will continue expanding this project to meet higher standards and more demanding scenarios. 
{\small
\bibliographystyle{ieee_fullname}
\bibliography{egbib}
}

\clearpage

\appendix

\section{Detailed Latency and Throughput Benchmark}\label{app:bench}
\subsection{Setup}
Unless otherwise stated, all the reported latency is measured on an NVIDIA Tesla T4 GPU with TensorRT version 7.2.1.6. Due to the large variance of the hardware and software settings, we re-measure latency and throughput of all the models under the same configuration (both hardware and software). For a handy reference, we also switch TensorRT versions (Table~\ref{tab:latency-qps-trt82}) for consistency check. Latency on a V100 GPU (Table~\ref{tab:latency-qps-v100}) is included for a convenient comparison. This gives us a full spectrum view of state-of-the-art detectors.

\subsection{T4 GPU Latency Table with TensorRT 8}

See \cref{tab:latency-qps-trt82}. The throughput of YOLOv6 models still emulates their peers.

\begin{table}[ht]
	\centering
	\resizebox{0.8\columnwidth}{!}{
		\begin{tabular}{l|HHHc|c|cHH}
			\toprule
			\multirow{2}{*}{\textbf{Method}} & \multirow{2}{*}{\textbf{Input}} & \multirow{2}{*}{\textbf{AP}$^{val}$} & \multirow{2}{*}{\textbf{AP}$_{50}^{val}$} &\multirow{2}{*}{\textbf{FPS}} &\multirow{2}{*}{\textbf{FPS}} &\multirow{2}{*}{\textbf{Latency}} & \multirow{2}{*}{\textbf{Params}} & \multirow{2}{*}{\textbf{FLOPs}} \\
			& & & & \tiny{(bs=1)} & \tiny{(bs=32)} &\tiny{(bs=1)} & & \\			
			\midrule
			\midrule
      YOLOv5-N~\cite{yolov5} & 640 & 28.0\% & 45.7\% & 702 & 843 & 1.4 ms& 1.9 M & 4.5 G\\
      YOLOv5-S~\cite{yolov5} & 640 & 37.4\% & 56.8\% & 433 & 515 & 2.3 ms& 7.2 M & 16.5 G \\
      YOLOv5-M~\cite{yolov5} & 640 & 45.4\% & 64.1\% & 202 & 235 & 4.9 ms& 21.2 M & 49.0 G  \\
      YOLOv5-L~\cite{yolov5} & 640 & 49.0\% & 67.3\% & 126  & 137 & 7.9 ms& 46.5 M & 109.1 G  \\
      \midrule
			\midrule
      YOLOX-Tiny~\cite{ge2021yolox} & 416 & 32.8\% & 50.3\%$^*$  & 766 & 1393 &1.3 ms & 5.1 M & 6.5 G \\
      YOLOX-S~\cite{ge2021yolox} & 640 & 40.5\% & 59.3\%$^*$ & 313  & 489 & 2.6 ms & 9.0 M & 26.8 G \\
      YOLOX-M~\cite{ge2021yolox} & 640 & 46.9\% & 65.6\%$^*$ & 159  & 204 & 5.3 ms &25.3 M & 73.8 G \\
      YOLOX-L~\cite{ge2021yolox} & 640 & 49.7\% & 68.0\%$^*$ &104  & 117 & 9.0 ms &54.2 M &155.6 G \\
      \midrule
      \midrule
      PPYOLOE-S~\cite{xu2022ppyoloe} & 640 & 43.1\% & 59.6\% & 357 & 493 & 2.8 ms& 7.9 M & 17.4 G \\
      PPYOLOE-M~\cite{xu2022ppyoloe} & 640 & 49.0\% & 65.9\% & 163 & 210 & 6.1 ms& 23.4 M& 49.9 G \\
      PPYOLOE-L~\cite{xu2022ppyoloe} & 640 & 51.4\% & 68.6\% & 110  & 145 & 9.1 ms& 52.2 M& 110.1 G \\
      \midrule
      \midrule
    %  YOLOv7-Tiny~\cite{wang2022yolov7} & 416 &33.3\%$^*$ & 49.9\%$^*$ & 787  &1196  & 1.3 ms & 6.2 M & 5.8 G \\
      YOLOv7-Tiny~\cite{wang2022yolov7} & 640 &37.4\%$^*$ & 55.2\%$^*$ & 464  &   568 & 2.1 ms & 6.2 M & 13.7 G$^*$ \\
      YOLOv7~\cite{wang2022yolov7}  & 640 & 51.2\% & 69.7\% & 128 & 135 & 7.6 ms & 36.9 M& 104.7 G \\
      \midrule
      \midrule
      YOLOv6-N & 640 & 35.9\% & 51.2\% & 810 & 1323 & 1.2 ms &4.3 M & 11.1 G \\
      YOLOv6-T & 640 & 40.3\% & 56.6\% & 469 & 677 & 2.1 ms & 15.0 M & 36.7 G \\
      YOLOv6-S & 640 & 43.5\% & 60.4\% & 362 & 522 & 2.7 ms & 17.2 M & 44.2 G \\
      YOLOv6-M & 640 & 49.5\% & 66.8\%& 180 & 241 & 5.6 ms & 34.3 M & 82.2 G \\
      YOLOv6-L-ReLU & 640 &  & & 111 & 145 & 9.0 ms & 34.3 M & 82.2 G \\
      YOLOv6-L & 640 & 52.3\% & 69.9\% & 105  & 131 & 9.6 ms & 58.4 M & 144.0 G \\
      \bottomrule
		\end{tabular}
	}
	\caption{
    YOLO-series comparison of latency and throughput on a T4 GPU with a higher version of TensorRT (8.2).
	}
	\label{tab:latency-qps-trt82}
\end{table}

\subsection{V100 GPU Latency Table}
See~\cref{tab:latency-qps-v100}. The speed advantage of YOLOv6 is largely maintained.

\begin{table}[ht]
	\centering
	\resizebox{0.8\columnwidth}{!}{
		\begin{tabular}{l|HHHc|c|cHH}
			\toprule
			\multirow{2}{*}{\textbf{Method}} & \multirow{2}{*}{\textbf{Input}} & \multirow{2}{*}{\textbf{AP}$^{val}$} & \multirow{2}{*}{\textbf{AP}$_{50}^{val}$} &\multirow{2}{*}{\textbf{FPS}} &\multirow{2}{*}{\textbf{FPS}} &\multirow{2}{*}{\textbf{Latency}} & \multirow{2}{*}{\textbf{Params}} & \multirow{2}{*}{\textbf{FLOPs}} \\
			& & & & \tiny{(bs=1)} & \tiny{(bs=32)} &\tiny{(bs=1)} & & \\			
			\midrule
			\midrule
      YOLOv5-N~\cite{yolov5} & 640 & 28.0\% & 45.7\% &709  & 1891 & 1.3 ms& 1.9 M & 4.5 G\\
      YOLOv5-S~\cite{yolov5} & 640 & 37.4\% & 56.8\% & 571 & 1354 & 1.6 ms& 7.2 M & 16.5 G \\
      YOLOv5-M~\cite{yolov5} & 640 & 45.4\% & 64.1\% & 332 & 685 & 2.9 ms& 21.2 M & 49.0 G  \\
      YOLOv5-L~\cite{yolov5} & 640 & 49.0\% & 67.3\% &215  & 426 & 4.5 ms& 46.5 M & 109.1 G  \\
      \midrule
			\midrule
      YOLOX-Tiny~\cite{ge2021yolox} & 416 & 32.8\% & 50.3\%$^*$  &739  & 3271 &1.3 ms & 5.1 M & 6.5 G \\
      YOLOX-S~\cite{ge2021yolox} & 640 & 40.5\% & 59.3\%$^*$ &515  & 1254 & 1.9 ms & 9.0 M & 26.8 G \\
      YOLOX-M~\cite{ge2021yolox} & 640 & 46.9\% & 65.6\%$^*$ &308  & 605 & 3.2 ms &25.3 M & 73.8 G \\
      YOLOX-L~\cite{ge2021yolox} & 640 & 49.7\% & 68.0\%$^*$ &189  & 370 & 5.2 ms &54.2 M &155.6 G \\
      \midrule
      \midrule
      PPYOLOE-S~\cite{xu2022ppyoloe} & 640 & 43.1\% & 59.6\% & 435 & 1117 & 2.2 ms& 7.9 M & 17.4 G \\
      PPYOLOE-M~\cite{xu2022ppyoloe} & 640 & 49.0\% & 65.9\% & 263 & 583 & 3.8 ms& 23.4 M& 49.9 G \\
      PPYOLOE-L~\cite{xu2022ppyoloe} & 640 & 51.4\% & 68.6\% & 194 & 415 & 5.1 ms& 52.2 M& 110.1 G \\
      \midrule
      \midrule
    %  YOLOv7-Tiny~\cite{wang2022yolov7} & 416 &33.3\%$^*$ & 49.9\%$^*$ & 787  &1196  & 1.3 ms & 6.2 M & 5.8 G \\
      YOLOv7-Tiny~\cite{wang2022yolov7} & 640 &37.4\%$^*$ & 55.2\%$^*$ & 647  & 1269 & 1.5 ms & 6.2 M & 13.7 G$^*$ \\
      YOLOv7~\cite{wang2022yolov7}  & 640 & 51.2\% & 69.7\% & 229 & 425 & 4.3 ms & 36.9 M& 104.7 G \\
      \midrule
      \midrule
      YOLOv6-N & 640 & 35.9\% & 51.2\% & 752 & 2031 &1.0 ms &4.3 M & 11.1 G \\
      YOLOv6-T & 640 & 40.3\% & 56.6\% & 604 & 1724 & 1.4 ms & 15.0 M & 36.7 G \\
      YOLOv6-S & 640 & 43.5\% & 60.4\% & 507 & 1546 &1.7 ms & 17.2 M & 44.2 G \\
      YOLOv6-M & 640 & 49.5\% & 66.8\%& 287 & 750 &3.4 ms & 34.3 M & 82.2 G \\
      YOLOv6-L-ReLU & 640 & 52.3\% & 69.9\% & 198  & 476 & 5.0 ms & 58.4 M & 144.0 G \\
      YOLOv6-L & 640 & 52.3\% & 69.9\% & 175  & 416 & 5.6 ms & 58.4 M & 144.0 G \\
      \bottomrule
		\end{tabular}
	}
	\caption{
    YOLO-series comparison of latency and throughput on a V100 GPU. We measure all models at FP16-precision with the input size 640$\times$640 in the exact same environment.
	}
	\label{tab:latency-qps-v100}
\end{table}

\subsection{CPU Latency}
We evaluate the performance of our models and other competitors on a 2.6 GHz  Intel Core i7 CPU using OpenCV Deep Neural Network (DNN), as shown in~\cref{tab:latency-cpu}.

\begin{table}[ht]
	\centering
	\resizebox{0.7\columnwidth}{!}{
		\begin{tabular}{l|cHHHHcHH}
			\toprule
			\multirow{2}{*}{\textbf{Method}} & \multirow{2}{*}{\textbf{Input}} & \multirow{2}{*}{\textbf{AP}$^{val}$} & \multirow{2}{*}{\textbf{AP}$_{50}^{val}$} &\multirow{2}{*}{\textbf{FPS}} &\multirow{2}{*}{\textbf{FPS}} &\multirow{2}{*}{\textbf{Latency}} & \multirow{2}{*}{\textbf{Params}} & \multirow{2}{*}{\textbf{FLOPs}} \\
			& & & & \tiny{(bs=1)} & \tiny{(bs=32)} &\tiny{(bs=1)} & & \\			
			\midrule
			\midrule
      YOLOv5-N~\cite{yolov5} & 640 & 28.0\% & 45.7\% &691  & 1199 & 118.9 ms& 1.9 M & 4.5 G\\
      YOLOv5-S~\cite{yolov5} & 640 & 37.4\% & 56.8\% & 552 & 1054 & 202.2 ms& 7.2 M & 16.5 G \\
      \midrule
			\midrule
      YOLOX-Tiny~\cite{ge2021yolox} & 416 & 32.8\% & 50.3\%$^*$  &738  & 3074 &144.2 ms & 5.1 M & 6.5 G \\
      YOLOX-S~\cite{ge2021yolox} & 640 & 40.5\% & 59.3\%$^*$ &499  & 1223 & 164.6 ms & 9.0 M & 26.8 G \\
      YOLOX-M~\cite{ge2021yolox} & 640 & 46.9\% & 65.6\%$^*$ &295  & 598 & 357.9 ms &25.3 M & 73.8 G \\
      \midrule
      \midrule
    %  YOLOv7-Tiny~\cite{wang2022yolov7} & 416 &33.3\%$^*$ & 49.9\%$^*$ & 787  &1196  & 1.3 ms & 6.2 M & 5.8 G \\
      YOLOv7-Tiny~\cite{wang2022yolov7} & 640 &37.4\%$^*$ & 55.2\%$^*$ & 527  & 697 & 137.5 ms & 6.2 M & 13.7 G$^*$ \\
      \midrule
      \midrule
      YOLOv6-N & 640 & 35.9\% & 51.2\% & 752 & 2031 &70.0 ms &4.3 M & 11.1 G \\
      YOLOv6-T & 640 & 40.3\% & 56.6\% & 604 & 1724 & 128.1 ms & 15.0 M & 36.7 G \\
      YOLOv6-S & 640 & 43.5\% & 60.4\% & 507 & 1546 &163.4 ms & 17.2 M & 44.2 G \\
      \bottomrule
		\end{tabular}
	}
	\caption{
    YOLO-series comparison of latency on a typical CPU. We measure all models at FP32-precision with the input size 640$\times$640 in the exact same environment.
	}
	\label{tab:latency-cpu}
\end{table}

\section{Quantization Details}
\subsection{Feature Distribution Comparison}\label{app:quant-reopt-dist}
\begin{figure*}[ht]
    \centering
   \begin{subfigure}{0.45\textwidth}
   	 \includegraphics[width=0.9\columnwidth]{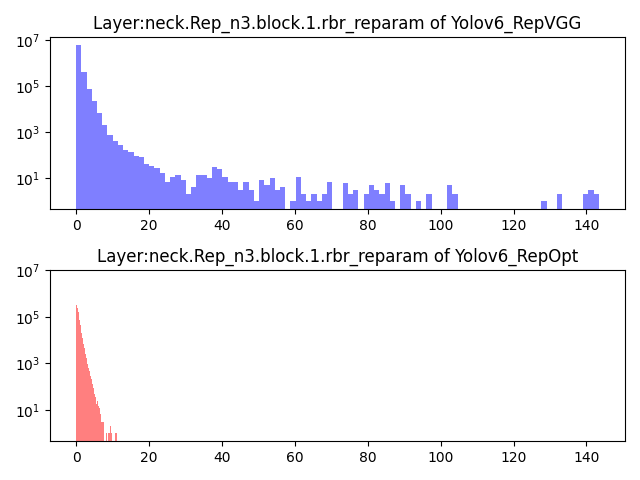}
	 \caption{}
    \end{subfigure}
     \begin{subfigure}{0.45\textwidth}
         \includegraphics[width=0.9\columnwidth]{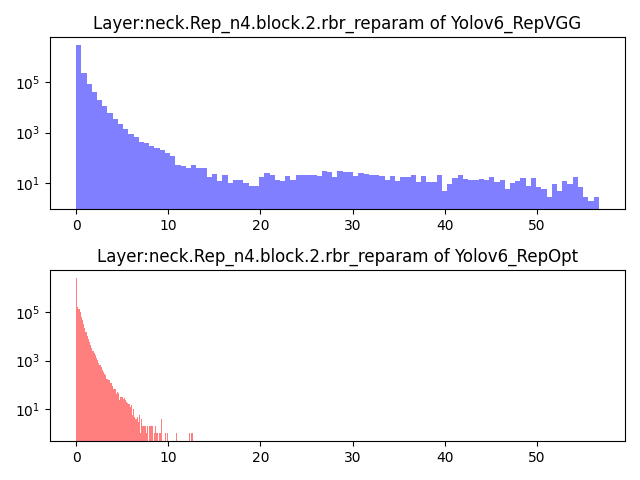}
         \caption{}
     \end{subfigure}
%     \begin{subfigure}{0.45\textwidth}
%    \includegraphics[width=0.9\columnwidth]{fig/40.png}
%    \caption{}
%   \end{subfigure}
%   \begin{subfigure}{0.45\textwidth}
%    \includegraphics[width=0.9\columnwidth]{fig/56.png}
%    \caption{}
%    \end{subfigure}
    \caption{More examples of better optimized layers in YOLOv6s that are otherwise hard to quantize.}
    \label{fig:more_repopt_act_map}
  \end{figure*}

We illustrate the feature distribution of more layers that are much alleviated after trained with RepOptimizer, see ~\cref{fig:more_repopt_act_map}.

\subsection{Sensitivity Analysis Results}\label{app:quant-sens}

\begin{figure*}[ht]
    \centering
    \includegraphics[clip, trim={0 0 0 1.2cm}, width=0.85\textwidth]{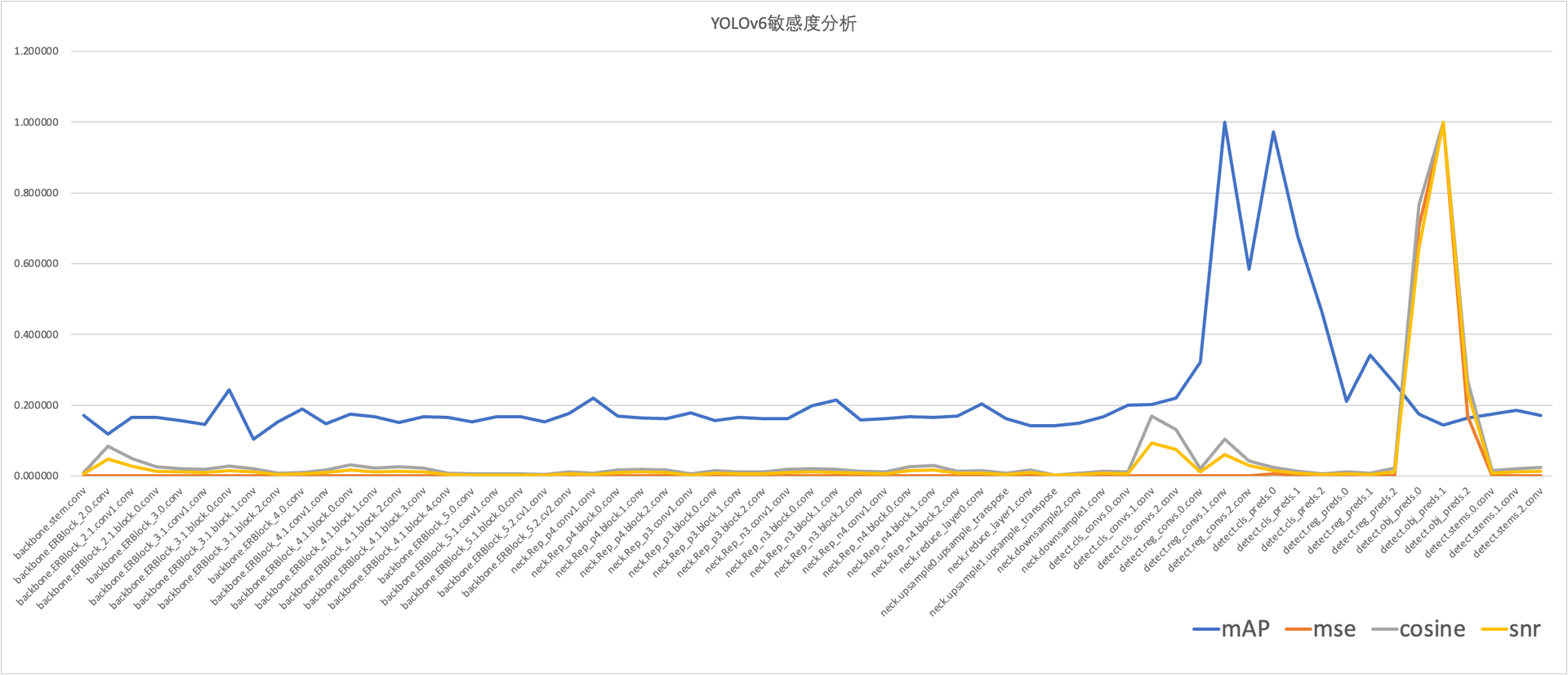}
    \caption{Quantization sensitivity analysis of all layers in YOLOv6s trained with RepOptimizer.}
    \label{fig:quant-sens}
  \end{figure*}

See ~\cref{fig:quant-sens}, we observe that SNR and Cosine similarity gives highly correlated results. However, directly evaluating AP produces a different panorama. Nevertheless, in terms of final quantization performance, MSE is the closest to direct AP evaluation, see Table~\ref{tab:quant-sens-map}.

\begin{table}[ht]
\centering
  \begin{tabular}{@{}l|c@{}}
    \toprule
    Model & AP$^{val}$ \\
    \midrule
    \midrule
    MSE & 41.5\\
    Cosine Similarity & 41.1 \\
    SNR & 41.1 \\
    Direct AP Evaluation & 42.0 \\
    \bottomrule
  \end{tabular}
\caption{Partial post-training quantization performance w.r.t. difference sensitivity metrics.}
\label{tab:quant-sens-map}
\end{table}

\section{Analysis of Gray Border}\label{app:gray-border}
\begin{figure*}[ht]
  \centering
 \begin{subfigure}{0.3\textwidth}
    \includegraphics[width=0.9\columnwidth]{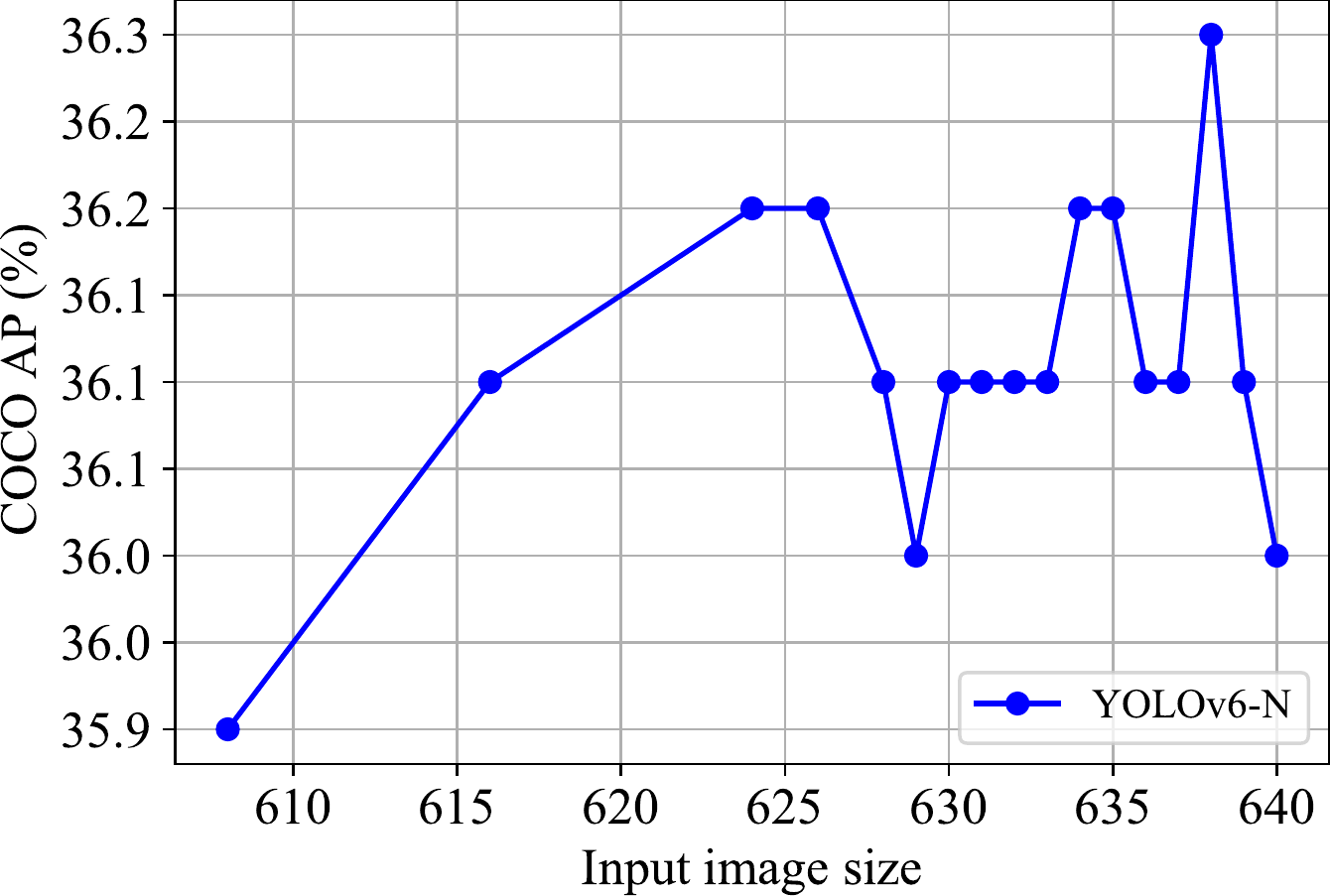}
 \caption{}
  \end{subfigure}
   \begin{subfigure}{0.3\textwidth}
       \includegraphics[width=0.9\columnwidth]{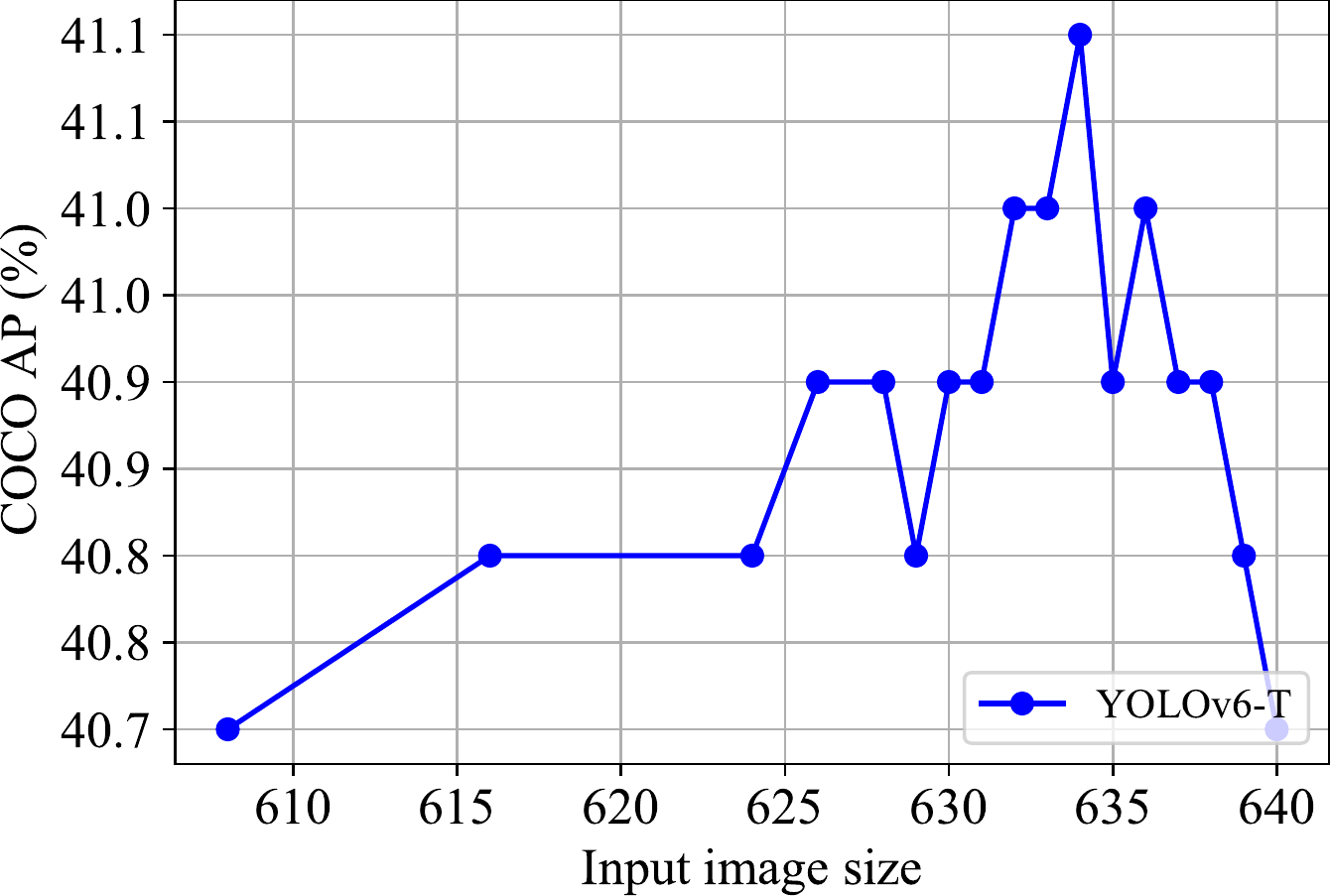}
       \caption{}
   \end{subfigure}
   \begin{subfigure}{0.3\textwidth}
  \includegraphics[width=0.9\columnwidth]{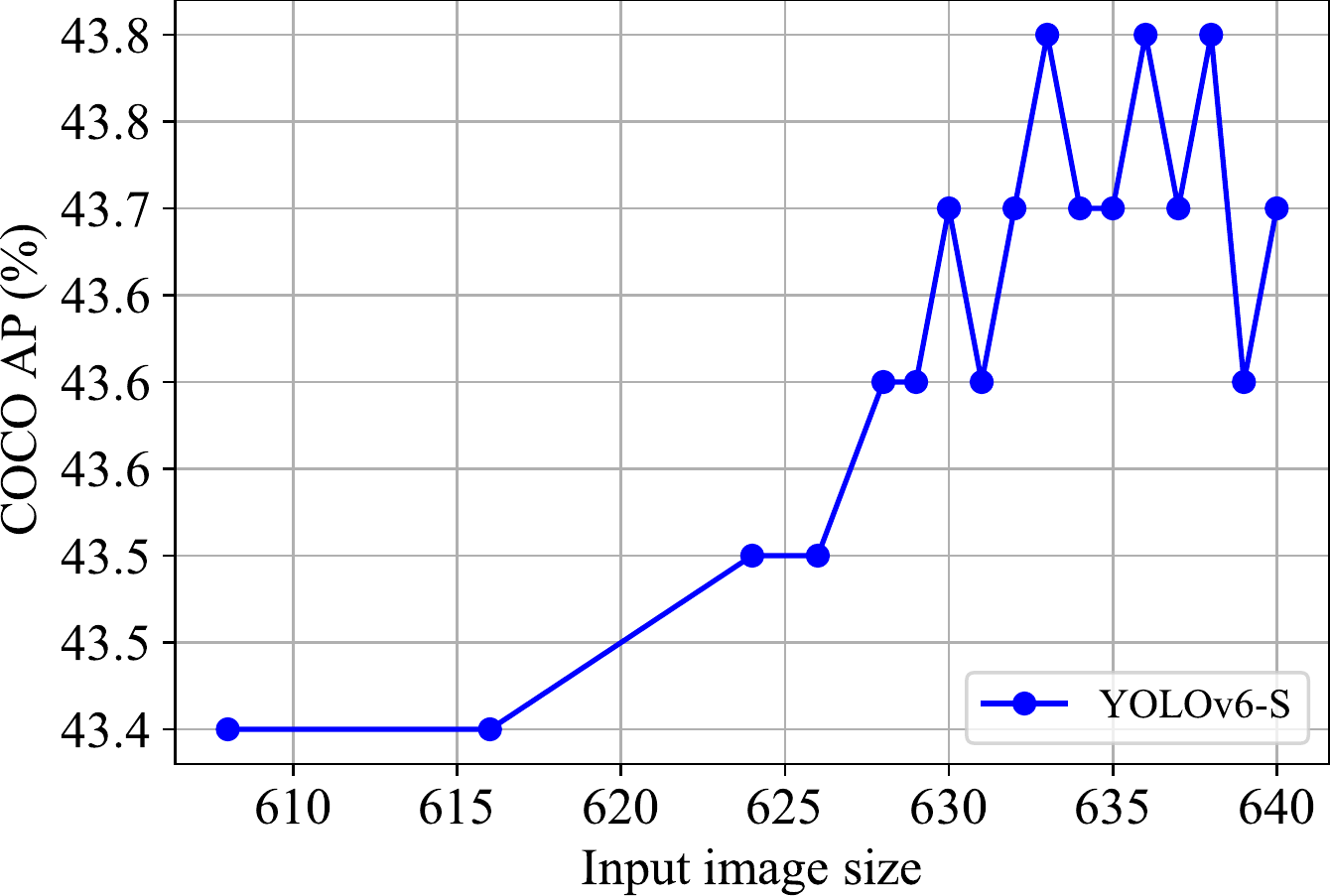}
  \caption{}
 \end{subfigure}
 \begin{subfigure}{0.3\textwidth}
  \includegraphics[width=0.9\columnwidth]{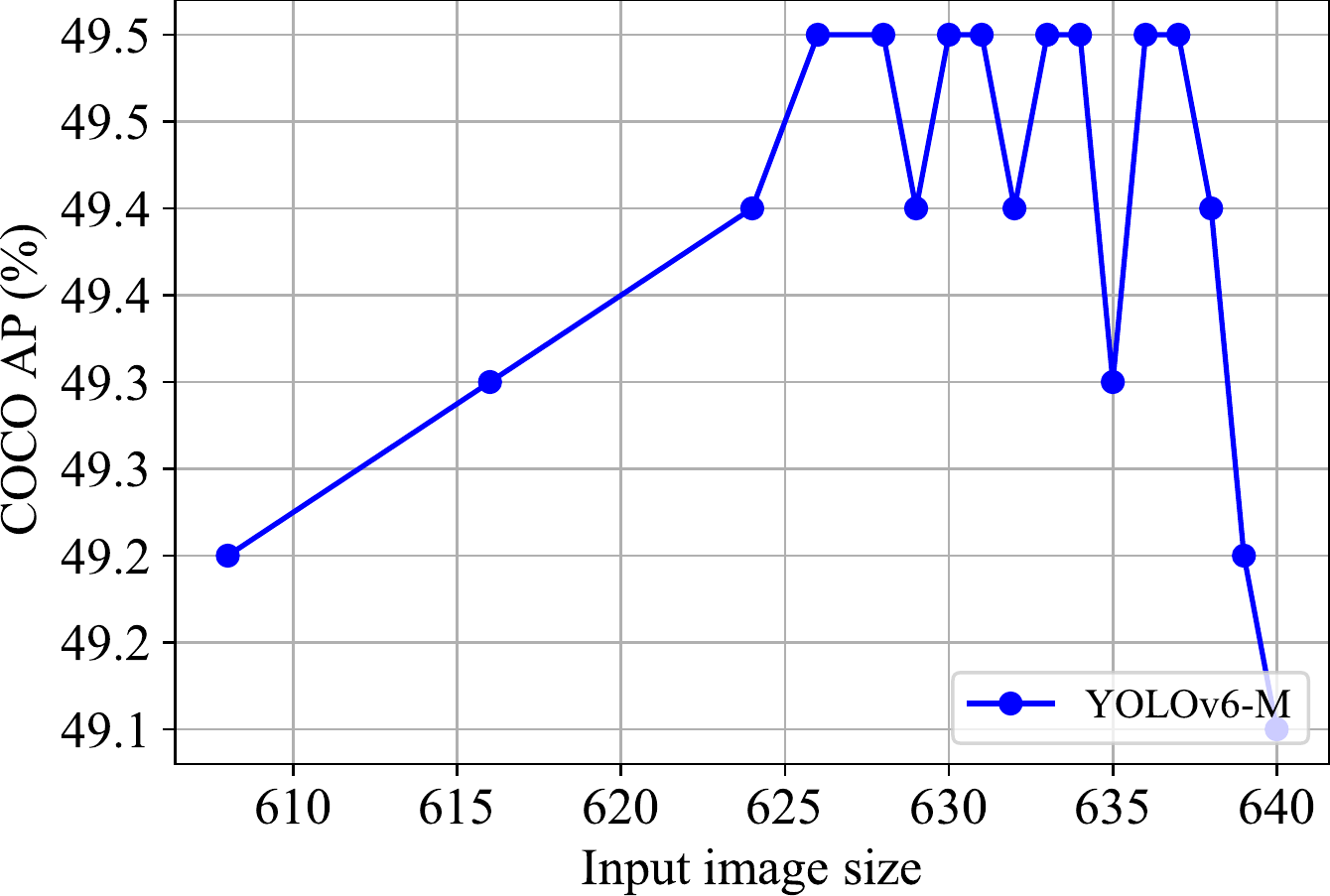}
  \caption{}
  \end{subfigure}
  \begin{subfigure}{0.3\textwidth}
    \includegraphics[width=0.9\columnwidth]{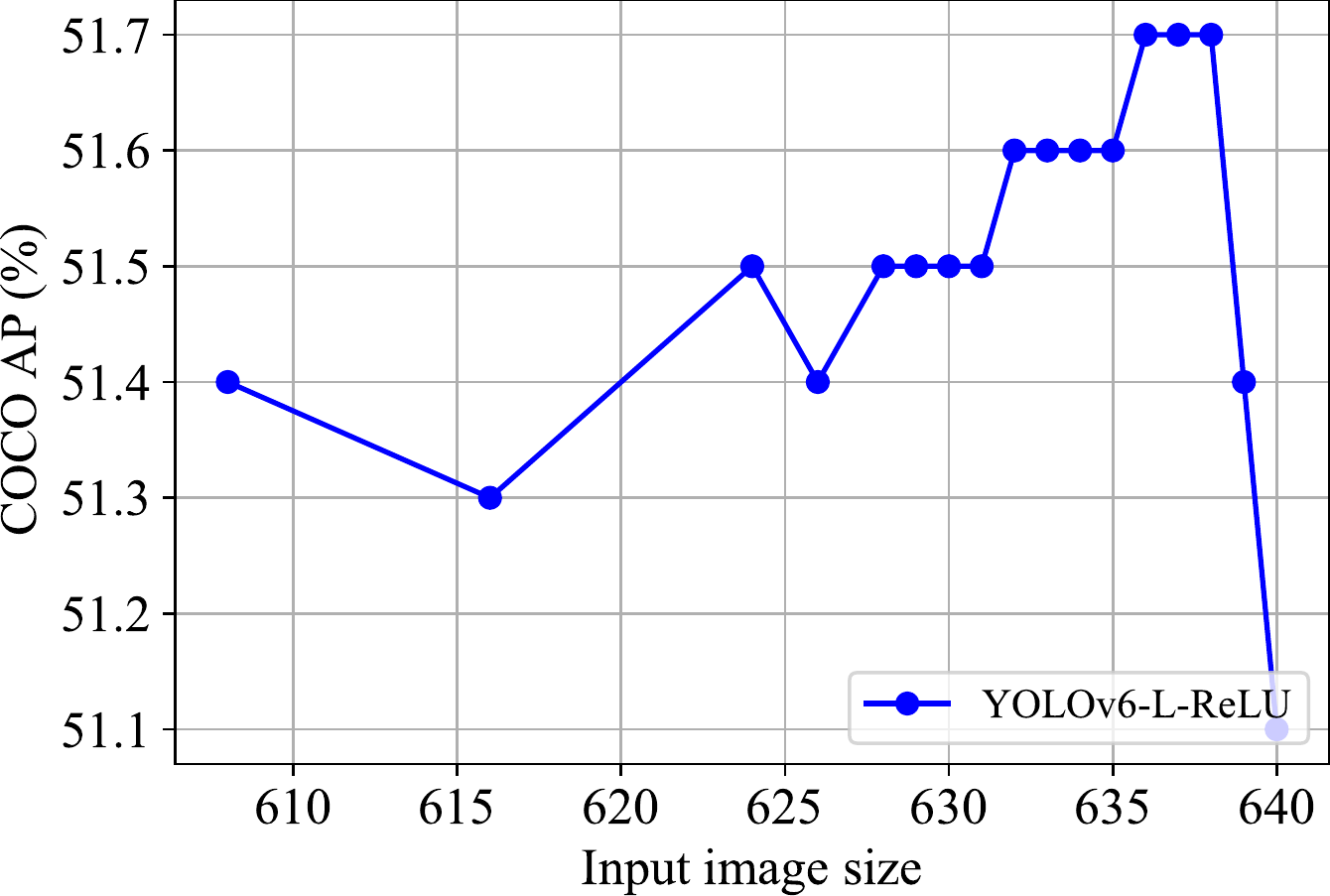}
    \caption{}
    \end{subfigure}
  \begin{subfigure}{0.3\textwidth}
    \includegraphics[width=0.9\columnwidth]{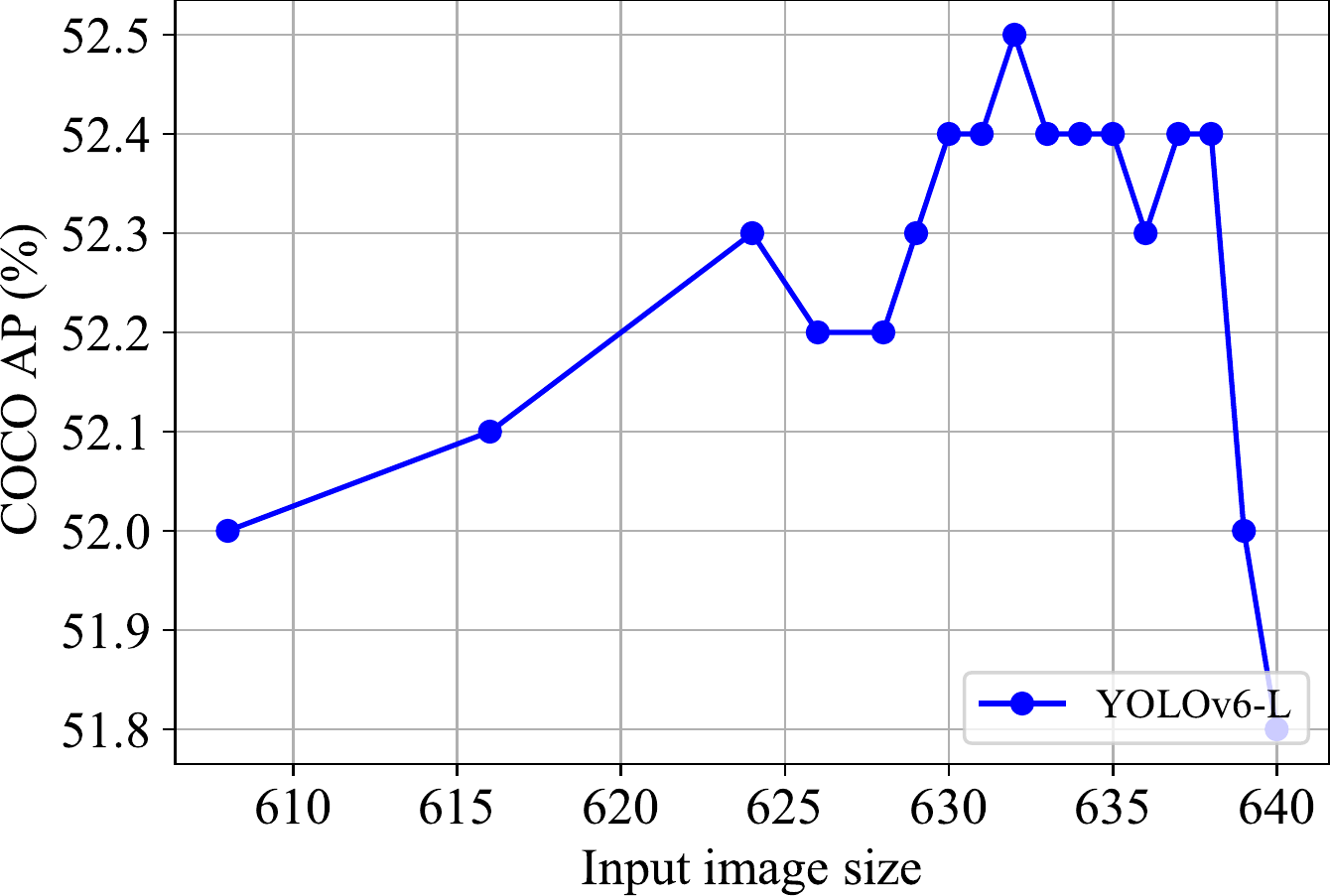}
    \caption{}
    \end{subfigure}
  \caption{Analysis of the gray border problem.}
  \label{fig:grayborder-appendix}
\end{figure*}

To analyze the effect of the gray border, we further explore different border settings with the loaded images resized to different sizes and padded to 640$\times$640. For example, when the image size is 608, and the border size is set to 16, and so on. In addition, we alleviate a problem of the information misalignment between pre-processing and post-processing via a simple adjustment. Results are shown in~\cref{fig:grayborder-appendix}. We can observe that each model achieves the best AP with a different border size. Additionally, compared with an input size of 640, our models get about 0.3\% higher AP on average if the input image size locates in the range from 632 to 638. 

\end{document}